\pdfoutput=1
\documentclass[11pt]{article}

\usepackage{acl}

\usepackage{times}
\usepackage{latexsym}
\usepackage{graphicx}
\usepackage{booktabs}
\usepackage{enumitem}
\usepackage{multirow}
\usepackage{microtype}
\usepackage{lipsum}
\usepackage{stmaryrd}
\usepackage{amsmath}
\usepackage{multirow}
\usepackage{bbm}
\usepackage{amssymb}
\usepackage{booktabs}
\usepackage{adjustbox}
\usepackage{xspace}
\usepackage{algpseudocode}
\usepackage{algorithm}
\usepackage{caption}
\usepackage{subcaption}
\usepackage{color, colortbl}
\usepackage{tablefootnote}
\usepackage{enumitem}  % More control
\usepackage{lipsum}
\usepackage{xspace}
\usepackage{bm}
\usepackage{colortbl}

\usepackage[T1]{fontenc}
\usepackage[utf8]{inputenc}

\setlist{topsep=1pt,itemsep=1pt,partopsep=1pt, parsep=1pt}

\title{Cendol: Open Instruction-tuned Generative\\Large Language Models for Indonesian Languages}

\newcommand{\equalsign}{\footnotemark[1]\hspace{0.1cm}}

\author{
Samuel Cahyawijaya$^{1,4*}$, Holy Lovenia$^{2,4}$\equalsign, Fajri Koto$^{3,4}\equalsign$, Rifki Afina Putri$^{4,5}\equalsign$, \\ \bf{Emmanuel Dave$^{4\dagger}$, Jhonson Lee$^{4\dagger}$, Nuur Shadieq$^{4\dagger}$, Wawan Cenggoro$^{4\dagger}$,} \\
\bf{Salsabil Maulana Akbar$^{4\dagger}$, Muhammad Ihza Mahendra$^{4\dagger}$, Dea Annisayanti Putri$^{4\dagger}$} \\
\bf{Bryan Wilie$^{1,4}$, Genta Indra Winata$^{6,4}$}, \bf{Alham Fikri Aji$^{3,4}$\equalsign},\\
\bf{Ayu Purwarianti$^{4,7,8}$, Pascale Fung$^1$} \\
$^1$HKUST \quad $^2$AI Singapore \quad $^3$MBZUAI \quad $^4$IndoNLP \quad $^5$KAIST \\
$^6$Bloomberg \quad $^7$Institut Teknologi Bandung \quad $^8$Prosa.ai \\
}

\begin{document}
\maketitle
\begin{abstract}

Large language models (LLMs) show remarkable human-like capability in various domains and languages.
% Nonetheless, there is a disparity in quality, especially in low-resource languages, such as Indonesian, making it ineffective and inefficient for these languages.
However, a notable quality gap arises in low-resource languages, e.g., Indonesian indigenous languages, rendering them ineffective and inefficient in such linguistic contexts.
% Addressing the quality gap, we introduce Cendol, a collection of Indonesian LLMs comprising both decoder-only and encoder-decoder architectures over various model sizes.
To bridge this quality gap, we introduce Cendol, a collection of Indonesian LLMs encompassing both decoder-only and encoder-decoder architectures across a range of model sizes.
% We showcase the effectiveness of Cendol on a wide range of tasks, achieving $\sim$20\% improvement as well as its generalization to local languages in Indonesia.
We highlight Cendol's effectiveness across a diverse array of tasks, attaining $\sim$20\% improvement, and demonstrate its capability to generalize to unseen tasks and indigenous languages of Indonesia.
% Cendol exhibits human-like capabilities, resulting in approximately 20\% performance improvement across diverse tasks.
% We also showcase our Cendol$^{chat}$ models which show significant improvement in terms of human favorability but still fall behind in capturing local knowledge and cultural values in Indonesia.
Furthermore, Cendol models showcase improved human favorability despite their limitations in capturing indigenous knowledge and cultural values in Indonesia. In addition, we discuss the shortcomings of parameter-efficient tunings, such as LoRA, for language adaptation. Alternatively, we propose the usage of vocabulary adaptation to enhance efficiency. Lastly, we evaluate the safety of Cendol and showcase that safety in pre-training in one language such as English is transferable to low-resource languages, such as Indonesian, even without RLHF and safety fine-tuning.\footnote{Cendol models are released under Apache 2.0 license and will be made publicly available upon acceptance.}
% \footnote{Cendol models are released under Apache 2.0 license and will be made publicly available on \url{https://huggingface.co/indonlp}.}

\end{abstract}

\section{Introduction}

% Indonesia & LLM
Indonesia is the fourth most populous country in the world, with around 280 million people spread across more than 17,000 islands within a humongous area of $\sim$2 million square kilometers. With such a large archipelago surrounding the country, digital services become immensely crucial, making Indonesia the fourth largest internet user in the world, with $\sim$220 million users.
% \footnote{\url{https://www.statista.com/statistics/262966/number-of-internet-users-in-selected-countries/}}. 
Despite the huge demand, the technology supporting Indonesian digital businesses still lags compared to other much smaller countries. One aspect that it still left behind is the access to state-of-the-art large language model (LLM) technology, such as ChatGPT~\cite{openai2023chatgpt} and GPT4~\cite{openai2023gpt4}. Although these LLMs support Indonesian and its local languages, these LLMs often have much weaker language representation for such low-resource and underrepresented languages~\cite{cahyawijaya2023nusawrites,cahyawijaya2023nusacrowd,asai2023buffet}.

% Problem with existing LLMs for Indonesia
The weak language representation in existing LLMs hurts their ability to generate responses in Indonesian and other underrepresented languages.
% Additionally, this also leads to inefficiency during inference, because of the vocabulary mismatch ensuing in a much longer sequence length when tokenizing sentences in these languages~\cite{ahia-etal-2023-languages}. 
This also leads to inefficiency during inference due to the vocabulary mismatch, hence texts in these languages are tokenized into much longer tokens~\cite{ahia-etal-2023-languages}. 
% Figure~\ref{fig:ineffective_and_inefficient} showcases this ineffectiveness and inefficiency problem of LLMs when processing low-resource languages, such as Indonesian. 
Additionally, these LLMs are more prone to safety issues, e.g., giving unsafe responses~\cite{wang2023all}, hallucinations~\cite{guerreiro2023hallu-mt,bang2023multitask}, and jailbreaking~\cite{yong2023lowresource,deng2023multilingual}.

\begin{figure}[!t]
    \centering
    \resizebox{\linewidth}{!}{
        \includegraphics[trim=0 0 0 0,clip]{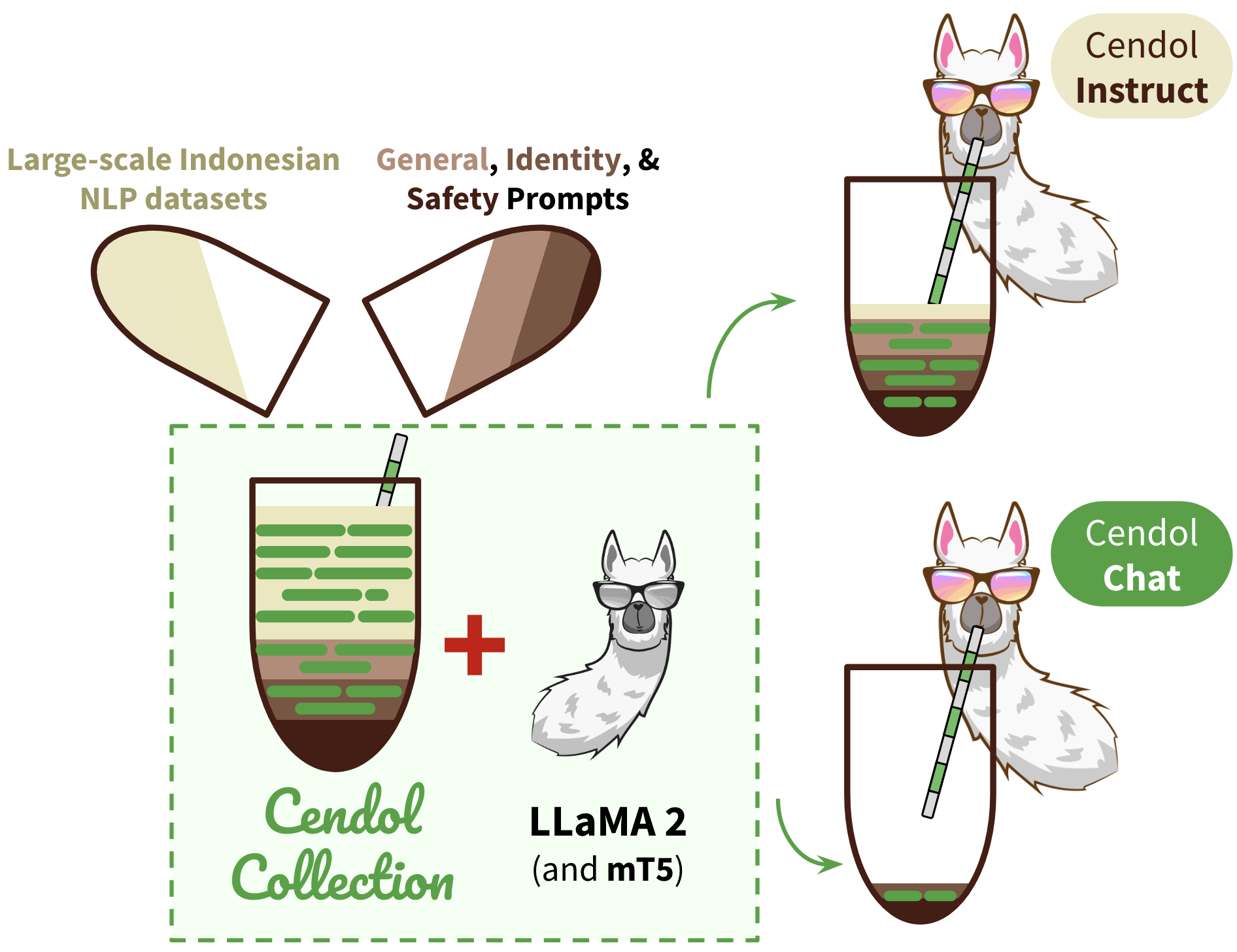}
    }
    \caption{Overview of Cendol Collection and LLM adaptation into Cendol$^{inst}$ and Cendol$^{chat}$ models.}
    \label{fig:cendol_overview}
\end{figure}

% Introducing Cendol
% \dummy{\lipsum[3]}
To overcome the challenge of weak language representation in Indonesian languages, we introduce Cendol\footnote{Cendol is an iced sweet dessert that contains droplets of pandan-flavored green rice flour jelly and coconut milk, served with palm sugar syrup. Cendol is popular across Southeast Asia, especially in Indonesia.}, a series of large-scale instruction-tuned LLMs specifically tailored for handling Indonesian indigenous languages. Cendol covers both decoder-only and encoder-decoder LLMs that spread across various scales from 300M up to 13B parameters. Various strategies are incorporated to enable instruction tuning across various scales. We assess the effectiveness of Cendol on a comprehensive evaluation suite, covering various general NLP tasks (e.g., sentiment analysis, topic modeling, machine translation, summarization, etc.), local knowledge, and cultural values evaluations.
% and evaluations involving safety and appropriateness of the generated responses.

% Contribution
Our work highlights the following contributions:
\begin{itemize}
    \item We introduce Cendol, a collection of state-of-the-art Indonesian LLMs, which outperforms all existing multilingual, Southeast Asian (SEA), and Indonesian LLMs.
    % and even significantly outperforms LLaMA2 with 70B parameters. 
    % Our largest Cendol variant shows comparable performance to commercial LLMs such as ChatGPT and GPT-4.
    \item We curate the Cendol Collection, a rigorous instruction-tuned corpus for Indonesian and local languages, covering 23 tasks and 10 languages, with a total of $\sim$50M instructions.
    \item We highlight the generalization of Cendol through a comprehensive evaluation suite, showcasing its adaptability towards various Indonesian NLP tasks and languages.
    % \item We show that improving the low-resource language representation, such as Indonesians, in existing LLMs can also improve the performance when using a high-resource language prompt while at the same time closing the performance gap between these languages. \holy{i'm not sure i fully understand what you want to deliver.}
    % \item We show the ineffectiveness of parameter-efficient tuning approaches, e.g., LoRA~\cite{hu2022lora}, on efficiently delivering high-quality regional LLMs. This raises the question of the importance of parameter-efficient methods for language adaptation.
    \item We demonstrate the ineffectiveness of parameter-efficient tuning approaches, exemplified by LoRA~\cite{hu2022lora}, in achieving high-quality regional LLMs. This prompts a consideration of the significance of parameter-efficient methods for language adaptation.
    \item We evaluate the safety of Cendol and showcase that safety in pre-training in one language such as English is transferable to low-resource languages, such as Indonesian.
    % \item We analyze the Safety and Efficiency aspect of regional LLM adaptation
\end{itemize}

\section{Related Work}

\subsection{Indonesian Language Models}
Various pre-trained Indonesian language models (LMs) have emerged in the past years, including IndoBERT~\cite{wilie-etal-2020-indonlu,koto-etal-2020-indolem,koto-etal-2021-indobertweet}, IndoBART~\cite{cahyawijaya-etal-2021-indonlg}, and IndoGPT~\cite{cahyawijaya-etal-2021-indonlg}. These models have smaller parameter sizes compared to recent LLMs and have primarily been evaluated only on standard NLP benchmarks. Concurrently, advancements in LLMs have led to the development of multilingual LLMs like BLOOM~\cite{scao2022bloom} and mT5~\cite{xue-etal-2021-mt5}, which include Indonesian, Javanese, and Sundanese.
% as well as local languages such as Javanese and Sundanese. 
Yet, they fall short of covering other underrepresented Indonesian local languages. LLaMA-2 \cite{touvron2023llama2}
% , a widely recognized open-source LLM, 
also incorporates Indonesian, although it comprises a small portion (0.03\%), diminishing its usability in the Indonesian context. Additionally, multilingual LLMs focusing on Southeast Asian languages, such as SEA-LION~\cite{sea_lion_2023} and SeaLLM~\cite{nguyen2023seallms}, are beginning to rise, indicating an increasing demand for refining LLMs for underrepresented languages.

\subsection{Instruction Tuning}
% \todo{fajri}

Instruction tuning is a technique to fine-tune LLMs using instruction-and-response pairs. Instruction-tuning allows zero-shot task generalization of LLMs 
% with enough variation of tasks and prompts
~\cite{sanh2022multitask,wei2022finetuned,ouyang2022training}. Various instruction-tuned LLMs have been developed, both monolingual and multilingual instruction-tuned LLMs using various backbone LLMs such as T5~\cite{raffel2020t5}, mT5~\cite{xue-etal-2021-mt5}, GPT-3~\cite{brown2020language}, BLOOM~\cite{scao2022bloom}, LLaMA~\cite{touvron2023llama}, LLaMA2~\cite{touvron2023llama2}, etc. Various efforts have created large-scale instruction-tuned datasets covering different types of instructions, including NLP task-specific instructions~\cite{sanh2022multitask,wei2022finetuned,muennighoff2022crosslingual,longpre2023flan,cahyawijaya-etal-2023-instructalign}, multi-turn conversation~\cite{wang2023openchat,vicuna2023}, safety prompts~\cite{bai2022constitutional,touvron2023llama2}, chain-of-thought instructions~\cite{wei2022cot,kojima2022cot,liu2023logicot}, etc. 

To better align with human preferences, instruction-tuning can also be coupled with reinforcement learning (RL). There are mainly two approaches for such alignment, i.e., reinforcement learning with human feedback (RLHF)~\cite{christiano2017rlhf,openai2023chatgpt} and reinforcement learning with artificial intelligence feedback (RLAIF)~\cite{lee2023rlaif,bai2022constitutional}. The reward models are trained to reflect human-preferred qualities such that RL enables the generated responses of LLMs to be more human-aligned. 
% In this work, we only apply supervised instruction-tuning for our Cendol models.

\subsection{LLM Evaluation in Indonesian Languages}
% \todo{bryan}
% \dummy{\lipsum[1]}
% In advancing LLM accessibility in Indonesia, significant efforts focus on evaluating their performance in Indonesian due to documented performance disparities across languages~\cite{koto-etal-2020-indolem,blasi2022systematic}. 
Due to the disparity of LLM performance across languages~\cite{blasi2022systematic}, significant efforts have focused on evaluating LLMs in Indonesian~\cite{koto-etal-2023-large,blasi2022systematic}. 
~\citet{scao2022bloom} evaluate BLOOM capabilities in Indonesian through slot-filling, intent classification, dialogue system,
% virtual assistant evaluation, 
and machine translation tasks. 
% BLOOM~\cite{workshop2022bloom} capabilities in Indonesian are evaluated through NLU tasks of slot-filling, intent classification, and virtual assistant evaluation in MASSIVE~\cite{fitzgerald-etal-2023-massive} and machine translation tasks in FLORES-101~\cite{goyal2022flores}. 
\citet{wei2023polylm, ahuja2023mega, asai2023buffet} compile multilingual NLU and NLG benchmarks and evaluated suites of LLMs in a wide range of NLP tasks.
% key tasks in Indonesian such as POS-tagging, NER, QA, and summarization.
% \citeauthor{wei2023polylm, ahuja2023mega, asai2023buffet} have made significant contributions by compiling multilingual NLU and NLG datasets to construct a multilingual benchmark addressing key tasks in Indonesian such as part of speech tagging in UDPOS~\cite{nivre2018universal}, named entity recognition in PAN-X~\cite{pan2017cross}, question-answering task in TydiQA~\cite{clark2020tydi}, summarization task in XLSum~\cite{hasan2021xl}, and commonsense reasoning tasks in XstoryCloze~\cite{lin2022few} and XCOPA~\cite{ponti2020xcopa}.
% \citeauthor{zhao2024llama} translated the LLM-Eval's instruction-following benchmark into Indonesian to explore the language capability transfer in Llama.

\begin{figure}[!t]
    \centering
    \resizebox{0.75\linewidth}{!}{
        \includegraphics[trim=0 0 0 0,width=0.7\textwidth,clip]{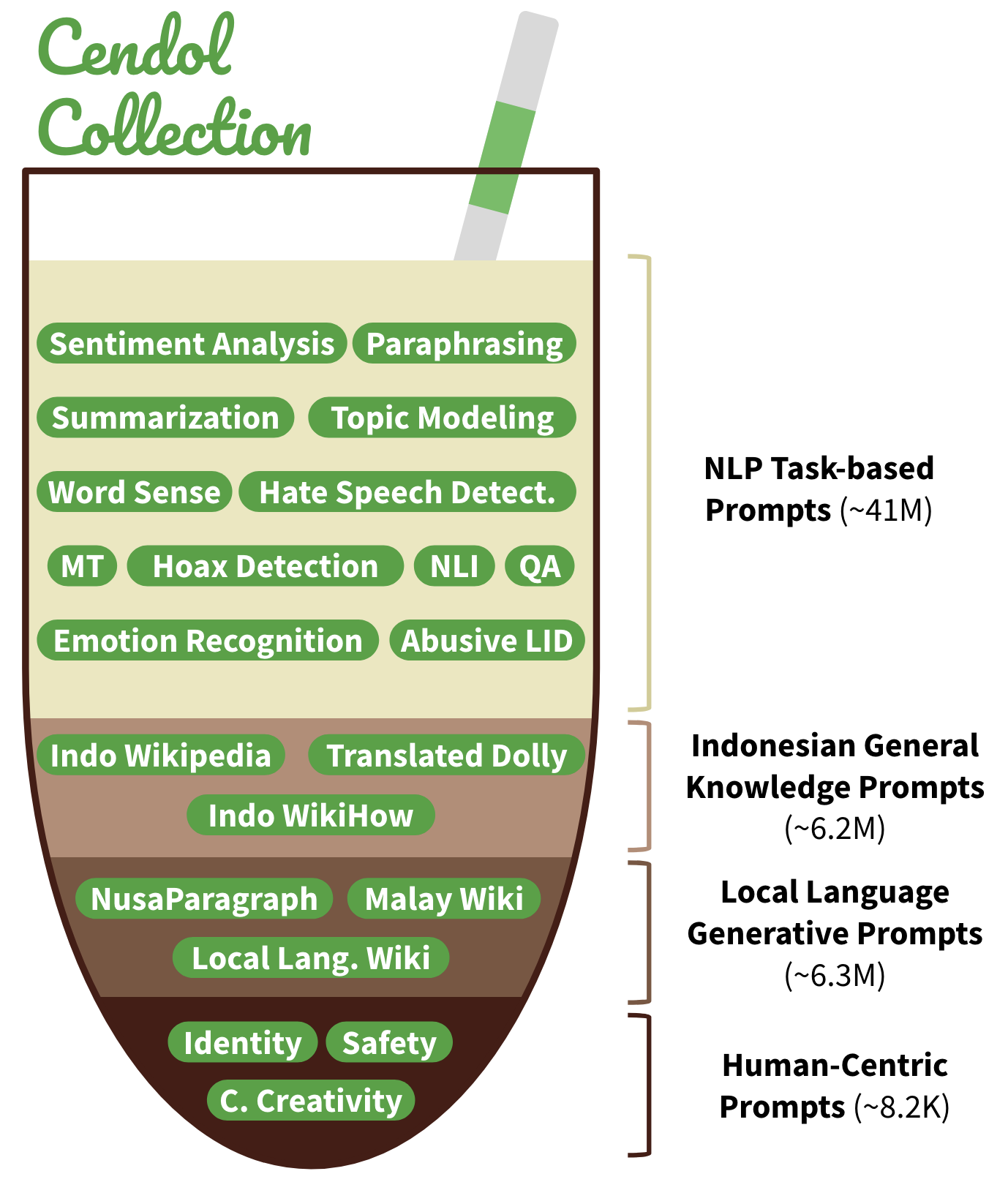}
    }
    \caption{The overview of Cendol Collection. Cendol Collection covers diverse prompts covering various types of instructions with a total of $\sim$53.5M prompts.}
    \label{fig:cendol-collection-overview}
\end{figure}

Further, \citet{koto-etal-2023-large, nguyen2023seallms} provide a more localized perspective by evaluating a suite of LLMs on multi-task language understanding benchmark for Indonesian culture and languages through questions from primary school to university entrance exams in Indonesia and M3Exam~\cite{zhang2023m3exam}. ChatGPT model family has also been put to the test by \citet{bang2023multitask, leong2023bhasa} for Indonesian NLU, NLG, and reasoning tasks. These evaluations highlight the challenges and opportunities in enhancing LLMs performance for Indonesian and its local languages, particularly addressing issues of dialectal variations and cultural context. We extend this research by evaluating the LLMs we develop across a spectrum of tasks, employing benchmarks specific to Indonesian indigenous languages.

\section{Cendol Collection}
% \todo{samuel}

In total, we create 53.5M prompts, covering  a wide range of prompt types including NLP task-based prompts (41M),  Indonesian general knowledge prompts (6.2M), local language generative prompts (6.3M), and human-aligned prompts (8.2K). Figure~\ref{fig:cendol-collection-overview} shows the detailed coverage of Cendol Collection across different sources.

%Cendol Collection covers a wide range of prompt types including NLP task-based prompts, 
% multi-turn prompt, 
%Indonesian general knowledge prompts, local language generative prompts, and human-aligned prompts. Figure~\ref{fig:cendol-collection-overview} shows the coverage of Cendol Collection.
% Each prompt type is described in the following sections.
% The complete list of datasets along with the metadata is shown in Appendix~\ref{app:cendol-collection}.

\subsection{NLP Task-Based Prompt}
\label{sec:nlp-prompt}

We collect NLP task-based prompts gathered from 124 dataset subsets covering various tasks, e.g., sentiment analysis, emotion recognition, topic modeling, hate speech detection,  natural language inference, machine translation, summarization, question answering, and paraphrasing. The datasets are gathered from NusaCrowd~\cite{cahyawijaya2023nusacrowd}. We gathered 10-20 prompts for each task type, resulting in a total of $\sim$41M prompts. 

% \subsection{Multi-Turn Prompt}
% We add multi-turn dialogue prompt XPersona

\subsection{Indonesian General Knowledge Prompt}
\label{sec:general-prompt}

To enable better generalization towards general knowledge, we extract general knowledge prompt from Indonesian Wikipedia\footnote{\url{https://id.wikipedia.org}} and Indonesian WikiHow.\footnote{\url{https://id.wikihow.com/}} 
% The prompt generation strategies for both Wikipedia and WikiHow are described in more detail in Appendix~\cite{app:general-prompt}. 
Additionally, we add an Indonesian machine-translated dataset from Databricks-Dolly-15k.\footnote{\url{https://huggingface.co/datasets/databricks/databricks-dolly-15k}} The dataset is translated using a distilled NLLB model with 1.3B parameters.\footnote{\url{facebook/nllb-200-distilled-1.3B}} In total, we accumulate $\sim$6.24M prompts for the Indonesian general knowledge prompt.

\subsection{Local Language Generative Prompt}
\label{sec:local-prompt}

To cover more underrepresented languages spoken in Indonesia, we collect Indonesian local language prompts from two sources, i.e., Wikipedia in local languages and NusaParagraph~\cite{cahyawijaya2023nusawrites}. We covered 18 local languages including Sundanese (sun), Javanese (jav), Acehnese (ace), Banjarese (bjn), Buginese (bug), and Gorontalo (gor). Since many local languages in Indonesia are derived from standard Malaysian (msa), we also collect the prompt from Malaysian Wikipedia.\footnote{\url{https://ms.wikipedia.org}} For the prompt from Wikipedia, we incorporate the same prompt generation strategy as in \S\ref{sec:general-prompt}, while for the generative prompt from NusaParagraph, we invert the input and output label of the dataset to make a sentence generation task for the specified local language. In total, we collect $\sim$6.27M local language generative prompts.

\subsection{Human-Centric Prompts}
\label{sec:human-prompt}
The quality of human-computer interaction is the essence of developing a dialogue agent. To improve the human-computer interaction quality of Cendol, we incorporate three types of human-centric prompts, i.e., identity prompt, safety prompt, and computational creativity prompt.

\paragraph{Identity Prompt}
Identity prompts are incorporated to provide a faithful identity of the Cendol models. These identity prompts include the personal identity of Cendol, the etymology of the word ``cendol'', the creator information of Cendol, and the neutrality of Cendol on various aspects, e.g., gender, religion, and political stance. In addition, we also include some trivia prompts to increase the engagingness of using Cendol. In total, we cover 125 identity prompts and to increase the representation of these prompts, we upsample the number of identity prompts by 500 in the Cendol Collection.

\paragraph{Safety Prompt}
We manually construct safety prompts to prevent Cendol from responding to queries that are not appropriate according to cultural norms and values in Indonesia. The safety prompts include prompts for guard-railing illegal activities, e.g., prostitution, gambling, illegal drugs, terrorism, racism, etc. Hate speech, offensive, and biased queries, especially regarding sensitive topics in Indonesia, such as religion and politics, are also guard-railed. In addition, we also prevent Cendol from providing unfaithful answers to queries that require knowledge from an expert, such as legal-related and medical-related queries. In total, we cover 187 safety prompts, and to increase the representation, we upsample the number of safety prompts to 500 in the Cendol Collection.

\paragraph{Computational Creativity}

Creativity is the essence of humanity~\cite{wilson2017origins}. To embed creativity into LLMs, we train Cendol with an open-source poem dataset, i.e., IndoPuisi~\cite{cahyawijaya2023nusacrowd}, endowing Cendol models the ability to generate Indonesian poems. The dataset covers 7,223 Indonesian poems and we upsample the number of the prompts by 20 in the Cendol Collection. 

\section{Cendol Recipe}

In this section, we describe the configurations for preparing our Cendol models and report the computational resources used in our experiments.

% \subsection{Tuning Strategy}

% Cendol models are instruction-tuned with the Cendol Collection dataset. 

\subsection{Backbone Models}

We prepare Cendol models of various base models to enable thorough comparison and analysis across different scales. Specifically, we train Cendol from models of different sizes, from 300M up to 13B parameters, to see the impact of size on performance. We also explore using decoder-only and encoder-only models, and lastly, we also explore using models of different origins to see if the starting base model has any impact. Finally, by producing Cendol in different configurations, users can choose their models based on needs and constraints. Specifically, we train Cendol by continuously fine-tuning decoder-only models, i.e., LLaMA-2 7B and LLaMA-2 13B~\cite{touvron2023llama2}, as well as encoder-decoder models, i.e., mT5$_{small}$, mT5$_{base}$, mT5$_{large}$, mT5$_{XL}$, mT5$_{XXL}$~\cite{xue-etal-2021-mt5}. For all backbone models with <10B parameters, we conduct a full parameter fine-tuning, while for >10B parameter models (i.e., LLaMA-2 13B and mT5$_{XXL}$), we utilize a parameter efficient fine-tuning approach, LoRA~\cite{hu2022lora}. 
% due to the limited computational resource available.

% \begin{table}[!t]
%     \centering
%     \resizebox{\linewidth}{!}{
%     \begin{tabular}{|c|c|c|c|c|}
%     \hline
%     \textbf{Eval Type} & \textbf{Task Type} & \textbf{Task} & \textbf{Lang. Type} & \textbf{Lang} \\ \hline
%     \multirow{12}{*}{NLU} & \multirow{8}{*}{Seen Task} & \begin{tabular}[c]{@{}c@{}}Entailment Classification\\ Sentiment Classification\\ Hate Speech Detection\end{tabular} & Seen Lang. & \begin{tabular}[c]{@{}c@{}}ind\end{tabular} \\ \cline{3-5} 
%      &  & Sentiment Analysis & Unseen Lang. & \begin{tabular}[c]{@{}c@{}}ace\\ ban\\ bjn\\ bug\\ mad\\ nij\end{tabular} \\ \cline{2-5} 
%      & Unseen Task & \begin{tabular}[c]{@{}c@{}}Answer Grading\\ Stance Detection\\ Next Tweet Prediction\\ Dialect Prediction\end{tabular} & Seen Lang. & \begin{tabular}[c]{@{}c@{}}ind\\ jav\end{tabular} \\ \hline
%     \multirow{5}{*}{NLG} & \multirow{5}{*}{Seen Task} & \multirow{2}{*}{Machine Translation} & Seen Lang. & \begin{tabular}[c]{@{}c@{}}sun-ind\\ jav-ind\\min-ind\\ bug-ind\\ bjn-ind\end{tabular} \\ \cline{4-5} 
%      &  &  & Unseen Lang. & \begin{tabular}[c]{@{}c@{}}ace-ind\\ ban-ind\end{tabular} \\ \cline{3-5} 
%      &  & \begin{tabular}[c]{@{}c@{}}Summarization\\ Paraphrasing\end{tabular} & Seen Lang. & \begin{tabular}[c]{@{}c@{}}ind\end{tabular} \\ \hline
%     \end{tabular}
%     }
%     \caption{Tasks and languages covered in our Indonesian and local languages benchmark evaluation suite.}
%     \label{tab:task_details}
% \end{table}

\begin{figure}[!t]
    \centering
    \resizebox{0.85\linewidth}{!}{
        \includegraphics[trim=0 0 0 0,width=0.7\textwidth,clip]{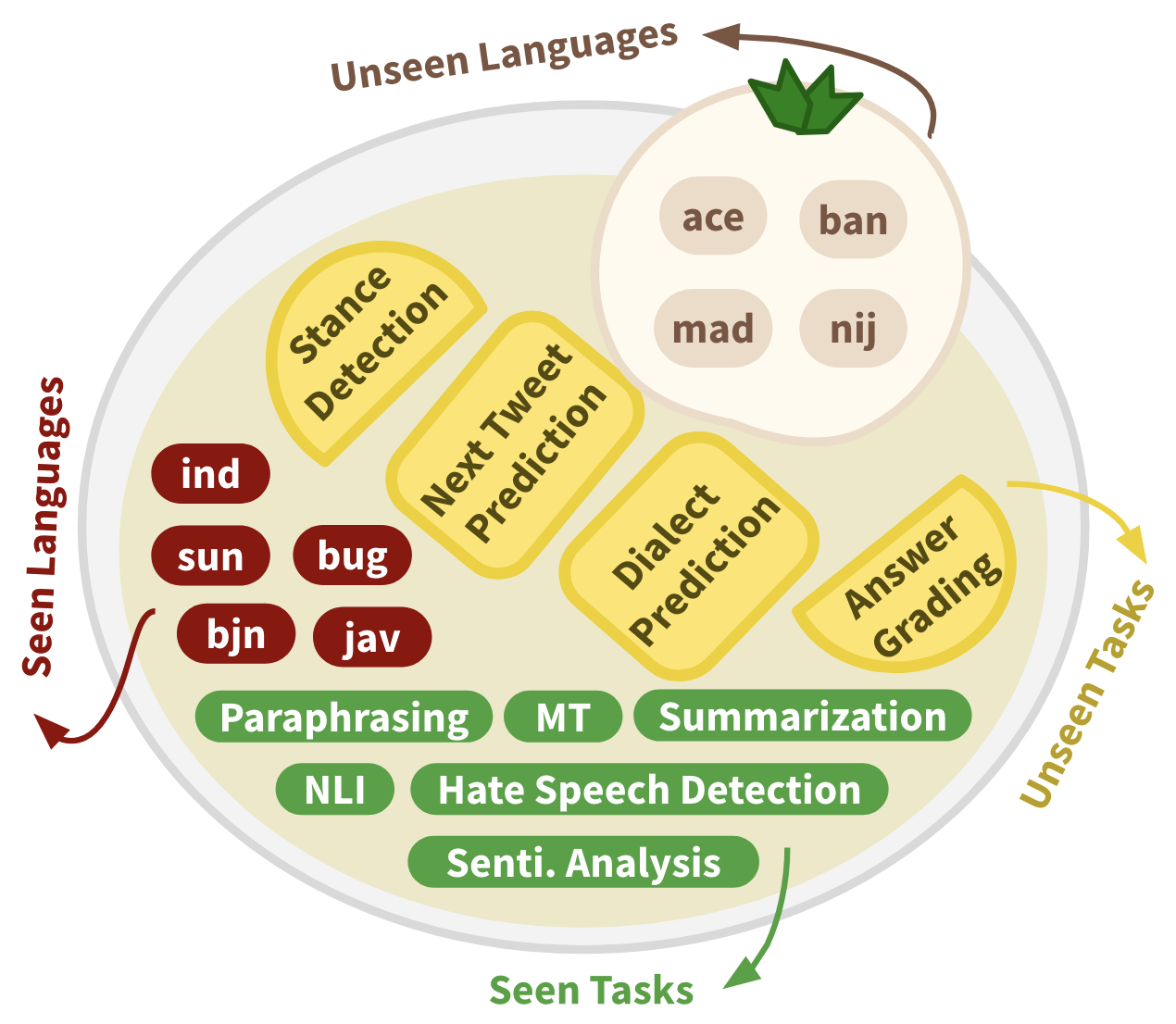}
    }
    \caption{Tasks and languages covered in our Indonesian indigenous benchmark evaluation suite.}
    \label{fig:tasks_details}
\end{figure}

% \paragraph{300M to 7B Models}
% \paragraph{13B Models}
% \todo{aji}
% \dummy{\lipsum[3]}

\begin{figure*}
    \centering
    \resizebox{0.975\linewidth}{!}{
        \includegraphics[trim=0 0.5em 0 0.5em,width=\textwidth,clip]{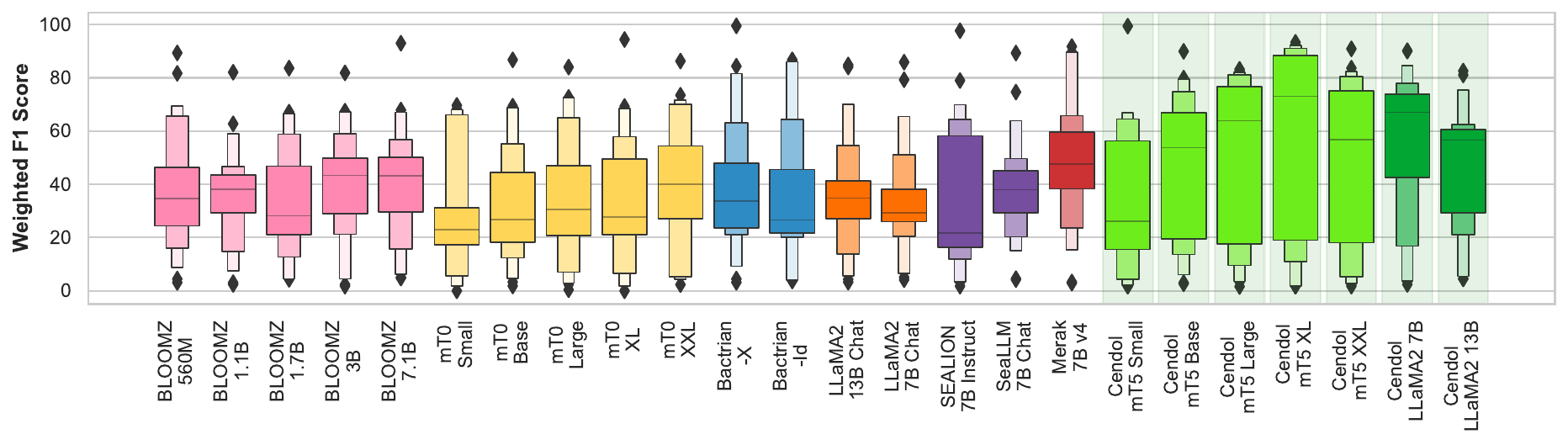}
    }
    \caption{Performance comparison of Cendol models with various multilingual, Southeast Asian, and Indonesian LLMs on NLU tasks. Largest fully fine-tuned Cendol variants, i.e, Cendol mT5$_{XL}$ and Cendol LLaMA2 7B, significantly outperform existing LLMs by $\sim$20\% weighted F1-score.}
    \label{fig:nlu-eval}
\end{figure*}

\subsection{Multi-Phase Tuning}
To develop a better instruction-tuned model, we develop the model in two phases of instruction-tuning for each backbone model. The first phase consists only of the NLP task-based prompt data with a total of 18 million instructions over-sampled from the NLP task-based prompt (\S\ref{sec:nlp-prompt}) in Cendol Collection. While the second phase consists of other prompt types including general knowledge prompts (\S\ref{sec:general-prompt}), local language generative prompts (\S\ref{sec:local-prompt}), and human-centric prompts (\S\ref{sec:human-prompt}) with a total of 12.8 million instructions. We divide the tuning into two phases to develop both stronger NLP task-specific and more general conversational LLMs. We denote the first phase models as Cendol-Instruct (Cendol$^{inst}$) and the second phase models as Cendol-Chat (Cendol$^{chat}$). We report the complete hyperparameters used in Appendix~\ref{app:hyperparameters}.

\subsection{Computational Resources} 
% \sam{Appendix}
% \todo{samcah and holy}

% \paragraph{Computational Resources}
% \todo{samcah and holy}
For the instruction tuning, we utilize a 4x40GB A100 GPU server for all models except for the fully fine-tuned LLaMA2-7B model where we use an 8x80GB A100 GPU server. We run the instruction-tuning using DeepSpeed ZeRO-3~\cite{rajbhandari2020zero} to optimize the computation time. The whole instruction tuning takes $\sim$40 days of training time, with around a 60:40 compute ratio between the first and the second phase instruction tuning. For evaluation, we run the evaluation on a single 40GB A100 GPU server.
% We showcase the carbon footprint of running each model on each phase in Appendix~\ref{app:carbon-footprint}.

\begin{figure*}
    \centering
    \begin{minipage}{0.91\linewidth}
        \resizebox{\linewidth}{!}{
            \includegraphics{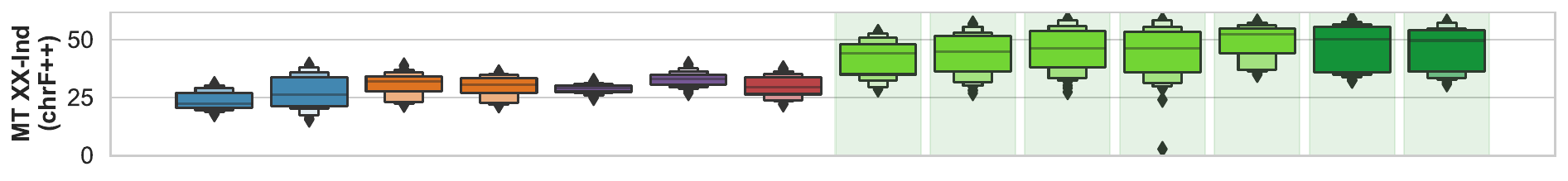}
        }
    \end{minipage}
    \begin{minipage}{0.91\linewidth}
        \resizebox{\linewidth}{!}{
            \includegraphics{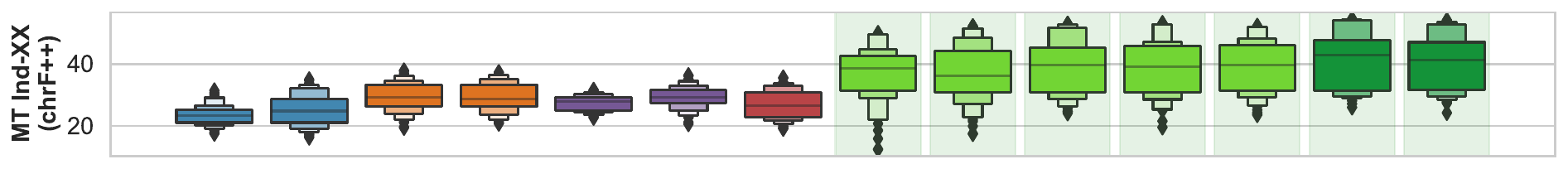}
        }
    \end{minipage}
    \begin{minipage}{0.91\linewidth}
        \resizebox{\linewidth}{!}{
            \includegraphics{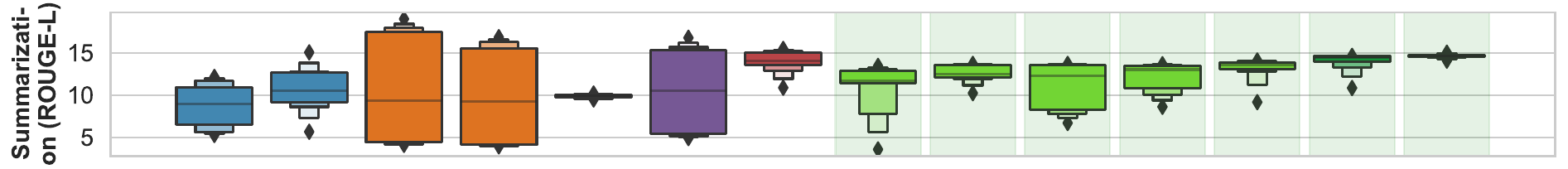}
        }        
    \end{minipage}
    \begin{minipage}{0.91\linewidth}
        \resizebox{\linewidth}{!}{
            \includegraphics{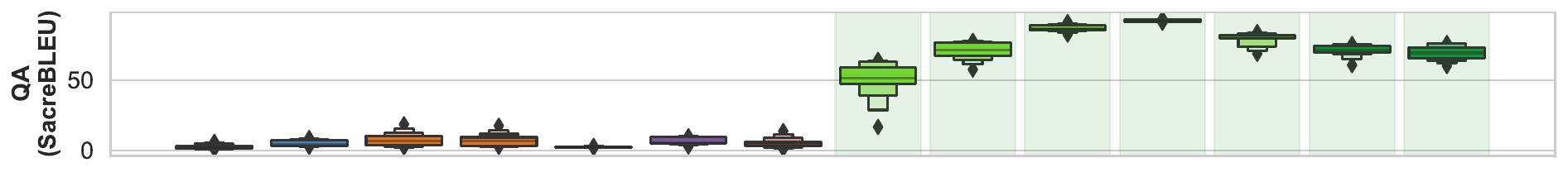}
        }        
    \end{minipage}
    \begin{minipage}{0.91\linewidth}
        \resizebox{\linewidth}{!}{
            \includegraphics[trim=0 0.75em 0 0,width=\textwidth,clip]{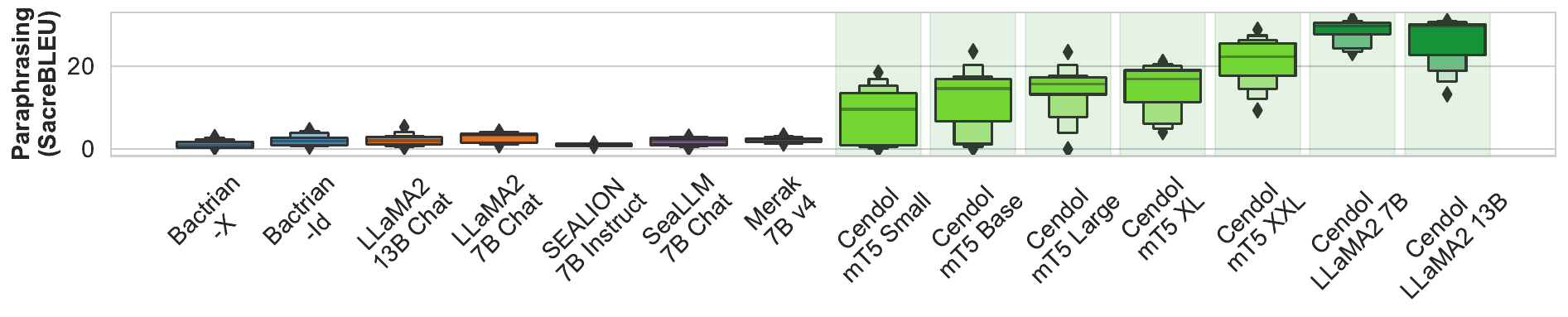}
        }        
    \end{minipage}
    \caption{Performance comparison of Cendol$^{inst}$ models with multilingual, SEA, and Indonesian LLMs on NLG tasks: \textbf{(1)} machine translation from local languages to Indonesian, \textbf{(2)} machine translation from Indonesian to local languages, \textbf{(3)} Indonesian language summarization, \textbf{(4)} Indonesian language question answering, and \textbf{(5)} Indonesian language paraphrasing. BLOOMZ and mT0 are not included since the evaluation datasets are exposed in xP3.}
    \label{fig:nlg-eval}
\end{figure*}

\section{Evaluation Suite}
% \todo{sabil, jhonson, dkk}

We evaluate Cendol on various aspects of language proficiency: NLU and NLG (\S\ref{sec:eval-nlp-nlg}); generalization capability on unseen tasks and languages (\S\ref{sec:eval-nlp-nlg}); as well as local knowledge and cultural commonsense ability (\S\ref{sec:eval-local-knowledge}). In addition, our evaluation includes the first Indonesian safety evaluation for LLMs (Appendix~\ref{app:safety}). Our evaluation suite consists of 10 local languages spoken in Indonesia and spreads across 23 evaluation datasets. 

\subsection{Indonesian Indigenous Evaluation}
\label{sec:eval-nlp-nlg}

To assess the language capability of Cendol models across Indonesian indigenous languages, we design an evaluation benchmark with 15 datasets covering 10 languages including Indonesian and 9 local languages spoken in Indonesia, i.e., Acehnese (ace), Balinese (ban), Banjarese (bjn), Buginese (bug), Javanese (jav), Madurese (mad), Minangkabau (min), Ngaju (nij), and Sundanese (sun).

As shown in Figure~\ref{fig:tasks_details}, this benchmark is split into four subsets: seen tasks, unseen tasks, seen languages, and unseen languages. The seen task subset shows how well the model performs on tasks it has encountered during training, while the unseen task subset assesses the model's ability to generalize to new tasks. The seen language subset and the unseen language subset test the model's performance and generalization to languages that are and are not part of the training data, respectively. For all tasks and datasets, we evaluate the model in a zero-shot prompting setting.

% Our benchmark is split into two main groups: \textbf{NLU tasks} for assessing more traditional NLP tasks, and \textbf{NLG tasks}, which include generating longer text passages. Within each group, we categorize the evaluation into three categories: \textbf{Seen Tasks}, \textbf{Unseen Tasks}, and \textbf{Unseen Languages}. Seen Tasks help us understand how well the model performs on tasks it encountered during training, while Unseen Tasks test the ability to generalize to new tasks. Unseen languages assess the performance and generalization to local languages that are not part of the training data. Unseen languages involve seen tasks like Machine Translation and Sentiment Analysis in six local languages. For all tasks and datasets, we evaluate the model in a zero-shot prompting setting. The details of each evaluation task are summarized in Figure~\ref{fig:tasks_details}.

\subsection{Local Knowledge and Cultural Commonsense Evaluation}
\label{sec:eval-local-knowledge}

Regional LLMs not only have to understand the local languages but also capture the understanding of local culture and nuances. To demonstrate this, we benchmark Cendol on several datasets. First, we test Cendol on the COPAL-ID~\cite{wibowo2023copal} dataset for local-nuanced commonsense reasoning. In COPAL-ID, a scenario is provided and two options are given, one of which is more plausible. All scenarios in COPAL-ID are infused with Indonesian local nuances and context. Next, we also utilize MABL~\cite{kabra-etal-2023-multi}, a binary classification dataset where the model is asked to interpret the meaning of a figure of speech in a sentence. We use the Indonesian, Javanese, and Sundanese subsets of MABL. We further benchmark Cendol on IndoStoryCloze~\cite{koto-etal-2022-cloze}, an Indonesian sentence completion dataset where the model is given two story endings, one of which is more plausible. Lastly, we also use MAPS~\cite{liu2023multilingual} that benchmarks LLM's understanding of multicultural proverbs and sayings.

\section{Impact and Consideration}
% \todo{Samcah compile results, Holy bikin gambar dan table}

\subsection{Comparison with Existing LLMs}
\label{sec:overall-benchmark}
% \todo{genta}
% \dummy{\lipsum[5-6]}

\begin{figure}
    \centering
    \resizebox{0.975\linewidth}{!}{
        \includegraphics[trim=0 0.5em 0 1em,width=\textwidth,clip]{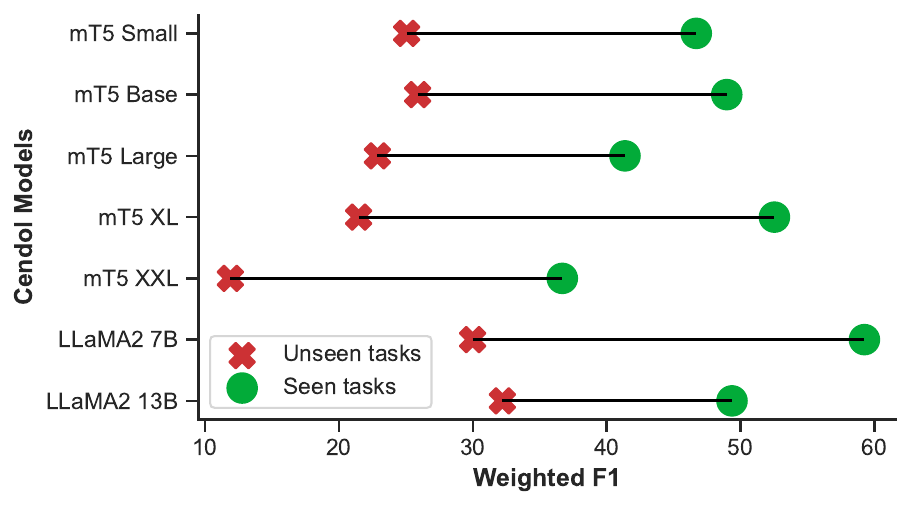}
    }
    \vspace{-5pt}
    \caption{Seen and unseen tasks performance of different Cendol models. All models consistently produce much lower performance for unseen tasks.}
    \label{fig:unseen-tasks}
\end{figure}

We present the results of the NLU and NLG evaluations of Cendol$^{inst}$ compared to existing LLMs in Figure~\ref{fig:nlu-eval} and Figure~\ref{fig:nlg-eval}, respectively. In terms of language understanding capability, the best Cendol$^{inst}$ model (i.e., Cendol mT5$_{XL}$) outperforms all existing LLMs both multilingual, SEA languages, and Indonesian LLMs on the comparable size by $\sim$20\% weighted F1-score. Even smaller Cendol$^{inst}$ models (i.e., Cendol mT5$_{base}$ and Cendol mT5$_{large}$ with 600M and 1.2B parameters, respectively), outperform larger LLMs with 7B and 13B parameters. Similarly for language generation, we observe huge improvements in MT, QA, and paraphrasing tasks with at least $\sim$20 increase in chrF++~\cite{popovic-2015-chrf} and SacreBLEU~\cite{post-2018-call}, respectively. For summarization tasks, Cendol$^{inst}$ models perform similarly to Merak 7B V4 and outperform other baseline LLMs by $\sim$5\% ROUGE-L. Our results signify the importance of large-scale instruction tuning to improve the zero-shot NLP capability for underrepresented regional languages.

\subsection{Generalization Towards Unseen Data}
\label{sec:unseen-generalization}

\begin{table}[!t]
    \centering
    \resizebox{0.975\linewidth}{!}{
    \begin{tabular}{c|c|c|c}
        \toprule
        \textbf{Model Type} & \textbf{Cendol}$\bm{^{inst}}$ & \textbf{Cendol}$\bm{^{chat}}$ & $\bm{\Delta}$\textbf{Perf}. \\
        \midrule
        Cendol mT5$_{small}$ & 30.02 & 29.84 & -0.18 \\
        Cendol mT5$_{base}$ & 45.08 & 35.87 & -9.21 \\
        Cendol mT5$_{large}$ & 48.82 & 40.13 & -8.69 \\
        Cendol mT5$_{XL}$ & 58.84 & 55.79 & -3.05 \\
        Cendol mT5$_{XXL}$ & 46.95 & 37.16 & -9.79 \\
        \midrule
        Cendol LLaMA2 7B & 56.80 & 50.34 & -6.46 \\
        Cendol LLaMA2 13B & 48.16 & 45.29 & -2.87 \\
        \bottomrule
    \end{tabular}
    }
    \caption{Comparison of NLU performance between Cendol$^{inst}$ and Cendol$^{chat}$ models. 
    % The difference is shown in $\Delta$Perf.
    }
    \label{tab:comparison_Instruct_Chat}
\end{table}

\begin{figure}
    \resizebox{0.9775\linewidth}{!}{
        \begin{minipage}{0.55\linewidth}
            \resizebox{\linewidth}{!}{
                \includegraphics[trim=0 4.25em 0 0,clip]{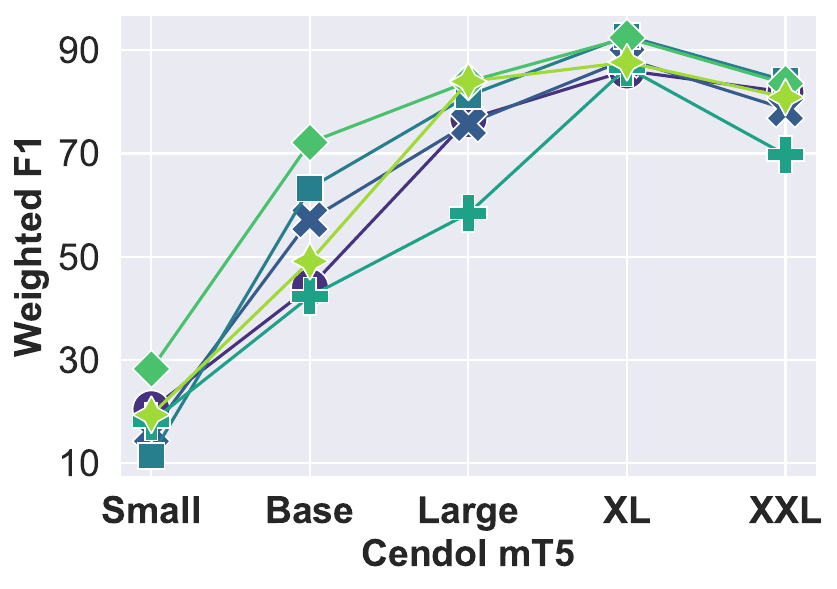}
            }
        \end{minipage}
        % \hspace{6pt}
        \begin{minipage}{0.35\linewidth}
            \resizebox{\linewidth}{!}{
                \includegraphics[trim=0 4.25em 0 0,clip]{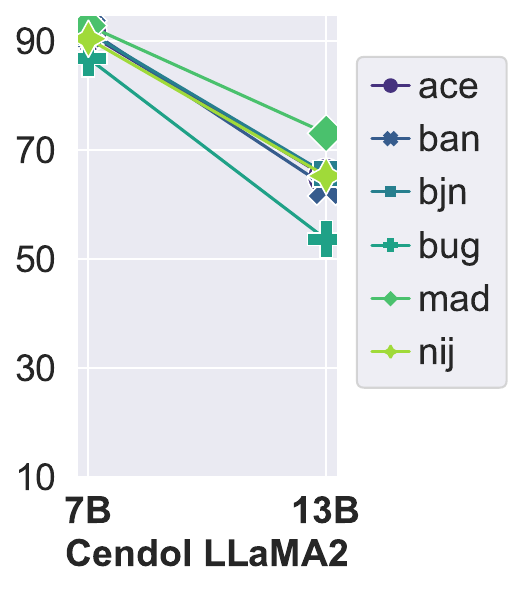}
            }
        \end{minipage}
    }
        % }
        % \hspace{5pt}
        % \fbox{
    \resizebox{\linewidth}{!}{
        \begin{minipage}{0.6\linewidth}
            \resizebox{\linewidth}{!}{
                \includegraphics{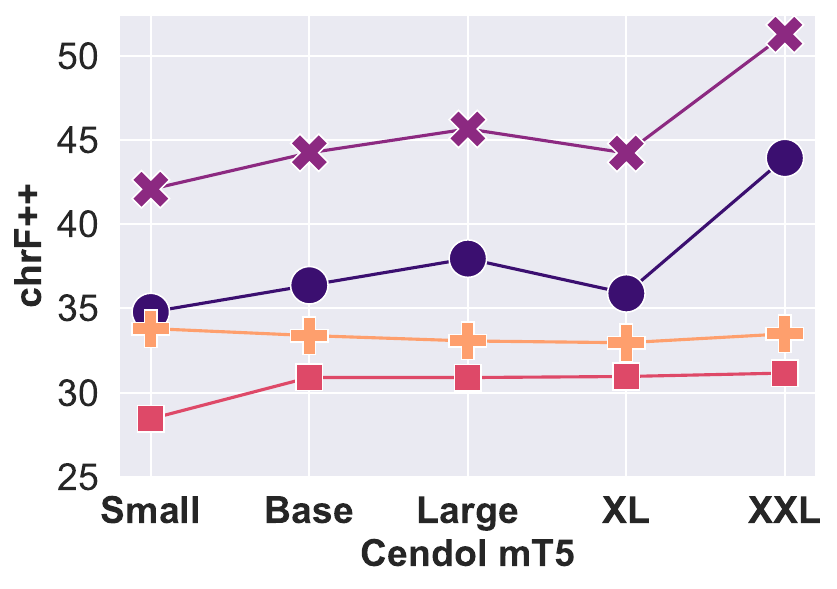}
            }
        \end{minipage}
        \begin{minipage}{0.4\linewidth}
            \resizebox{\linewidth}{!}{
                \includegraphics{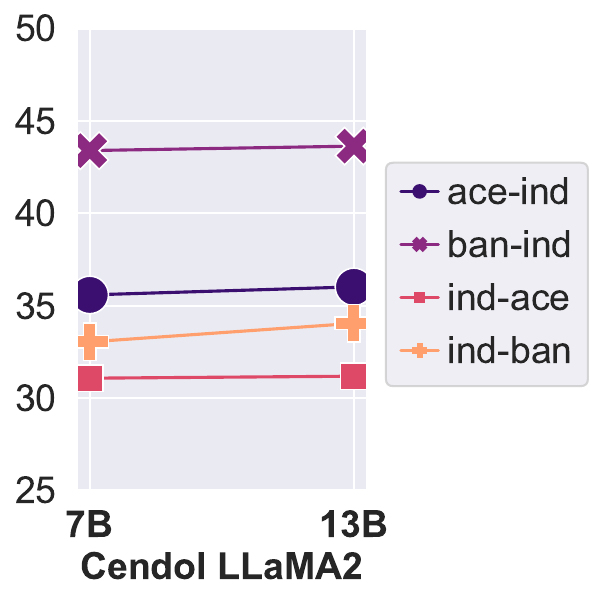}
            }
        \end{minipage}
    }
    % }
    \caption{Unseen language performance on \textbf{(top)} NLU and \textbf{(bottom)} NLG. Cendol shows considerable improvement as the model scales.
    % , while it is less obvious in NLG
    % , especially in the case of machine translation to unseen languages. 
    The drops on LLaMA2-13B and mT5$_{XXL}$ are due to the use of LoRA.}
    \label{fig:unseen-lang}
\end{figure}

\paragraph{Unseen Tasks} Figure~\ref{fig:unseen-tasks} showcases the performance of various Cendol models when evaluated on seen and unseen tasks. There is a huge performance drop (20\%-30\% weighted F1 score) in the unseen tasks, which can be attributed to two underlying reasons, i.e., 1) the difficulty of the unseen tasks themselves and 2) the generalization of the models towards the unseen tasks. When we fine-tune a smaller IndoBERT model~\cite{wilie-etal-2020-indonlu} into the seen and unseen tasks, there is a $\sim$10\% weighted F1 score difference between the unseen and seen tasks. Hence, we attribute the rest $\sim$10\%-20\% weighted F1 score as the generalization bottleneck of the Cendol models.

\paragraph{Unseen Languages} As shown in Figure~\ref{fig:unseen-lang}, the NLU performance to unseen languages follows the scaling law of LLMs. The performance improvements on NLG tasks are less apparent, moreover, no effect of scaling is observed on the translation from Indonesian to the unseen language direction. This showcases that, despite being able to better understand the unseen languages, the LLMs still have difficulty generating sentences in these unseen languages. Interestingly, we observe degradation in terms of NLU performance from the LoRA-tuned models, i.e., LLaMA2-13B and mT5$_{XXL}$, despite their increase in NLG performance. 
% We conjecture that this can happen because LoRA retains the embedding representation of the LLMs and we leave this exploration to future works.

\subsection{General vs. Task-Specific LLMs}
\label{sec:instruct-vs-chat}

\begin{figure}[!t]
    % \centering
    % \resizebox{0.95\linewidth}{!}{
    % \includegraphics[trim=0 3.2em 0 0, clip]{figures/human_eval/v1_eval.pdf}
    % }
    \centering
    \begin{minipage}{\linewidth}
        \resizebox{\linewidth}{!}{
            \includegraphics[trim=0 9.25em 0 0,clip]{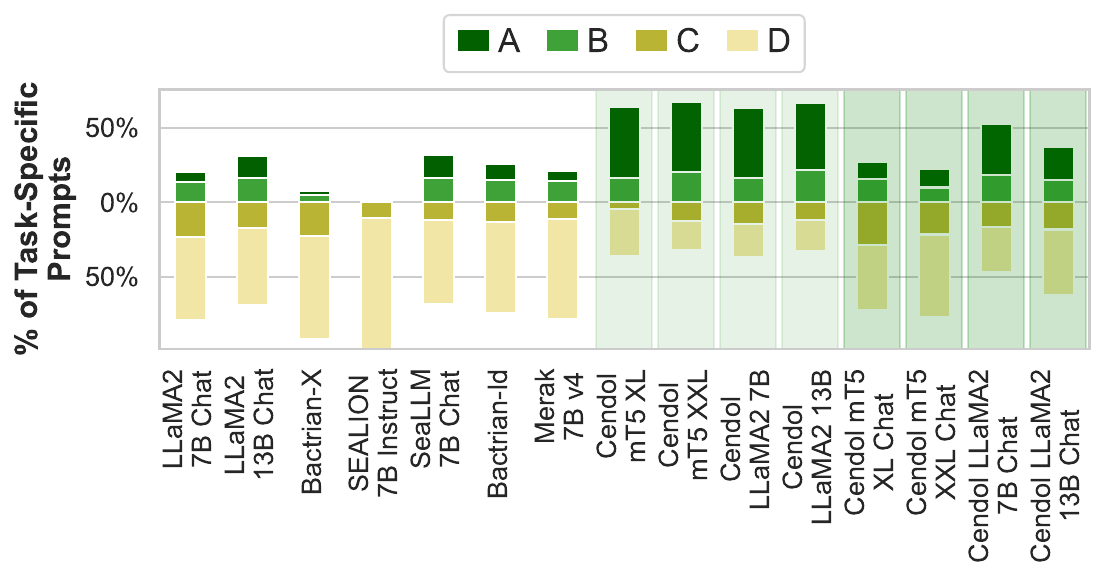}
        }
    \end{minipage}
    \begin{minipage}{\linewidth}
        \resizebox{\linewidth}{!}{
            \includegraphics[trim=0 0.5em 0 0,clip]{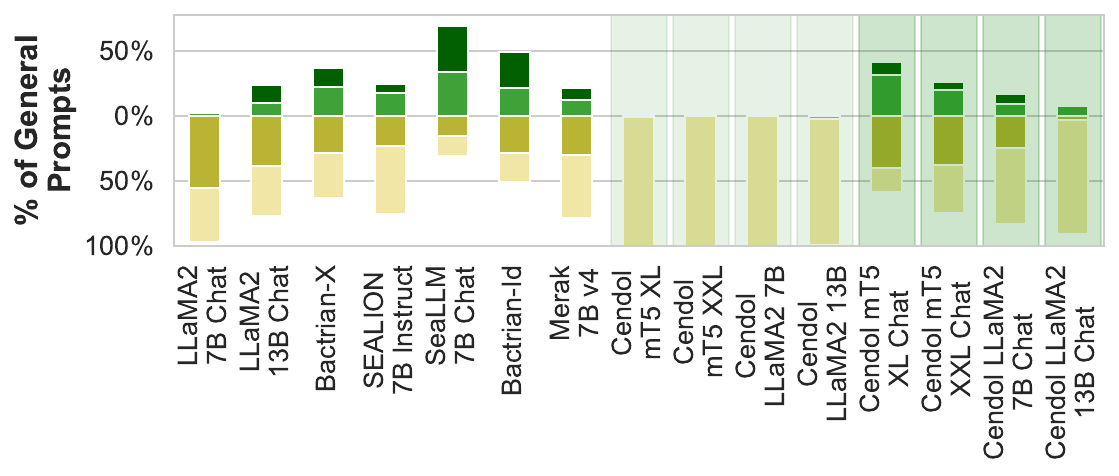}
        }
    \end{minipage}
    \caption{Human evaluation results of the baselines, Cendol$^{inst}$ models, and Cendol$^{chat}$ models on natural \textbf{(top)} task-specific and \textbf{(bottom)} general prompts prompts. A is the best and D is the worst.}
    \label{fig:human-eval}
\end{figure}

We compare the task-specific Cendol$^{inst}$ models with the general Cendol$^{chat}$ models. As shown in Table~\ref{tab:comparison_Instruct_Chat}, the task-specific performance of Cendol$^{chat}$ models decreases significantly by up to $\sim$10\% weighted F1-score. We further evaluate the models through human evaluation for both task-specific and general prompts (Figure~\ref{fig:human-eval}). For the task-specific human evaluation, we sample 60 generation results from all the evaluated NLG tasks. For the general prompts human evaluation, we generate responses from 100 prompts that require some local knowledge about Indonesia. The responses are then rated by 3 annotators with a moderate inter-annotator agreement ($\kappa$=0.59). The annotation guideline is described in Appendix~\ref{app:human-eval}. 

\paragraph{Task-specific prompts} The result of human evaluation on task-specific prompts is shown in Figure~\ref{fig:human-eval}. Cendol$^{inst}$ models significantly outperform all other models scoring a large portion of A rating compared to others. Cendol$^{chat}$ models achieve lower ratings, with LLaMA2-based Cendol$^{chat}$ performing slightly lower scores compared to the Cendol$^{inst}$ models, while mT5-based Cendol$^{chat}$ models perform on a par with other multilingual and regional LLMs.
% This showcases the effectiveness of transferring from 

\paragraph{General prompts} As shown in Figure~\ref{fig:human-eval}, Cendol$^{inst}$ models fail to answer general prompts in almost all cases, while Cendol$^{chat}$ models show a trend similar to the NLP task-specific performance of Cendol$^{inst}$ where larger models perform better than a smaller model with the exception on the LoRA-tuned model, i.e., mT5$_{XXL}$. Despite the huge quality shift after the second phase of instruction-tuning, there are only several responses that are accurate and comparable to human standards (Rate A). This shows that supervised fine-tuning alone is not enough and further human-alignment tuning strategy, such as RLHF~\cite{christiano2017rlhf} or RLAIF~\cite{bai2022constitutional}, is necessary to generate human-aligned responses.

% We futher compare the response quality of Cendol$^{chat}$ models with other LLMs. As shown in Figure~\ref{fig:xx}, Despite the general prompt instruction-tuning phase, the Cendol$^{chat}$ models are still outcompeted by existing LLMs such as LLaMA-2 Chat 7B and 13B models. We conjecture this might happen because of: 1) XXXX, 2) RLHF is more suitable than supervised fine-tuning for learning human-preferred responses for general prompts, and 3) ZZZZ.

\subsection{Capturing Local Knowledge}
\label{sec:local-knowledge}
% \todo{aji,fajri,samcah,genta}

\begin{table}[!t]
    \centering
    \resizebox{\linewidth}{!}{
        \begin{tabular}{lcccccccc}
        \toprule
        Model &   \multicolumn{3}{c}{MABL} &  MAPS &  COPAL & Indo & IndoStory \\
         & id &   jv &   su  & & & MMLU & Cloze \\
        \toprule
        \multicolumn{8}{c}{\textit{Multilingual LLM}} \\
        \midrule
        % BLOOMZ 560M & 54.10 & 52.21 & 51.50 & 40.95 & 51.84 & 17.96 & 55.14 \\
        % BLOOMZ 1.1B & 58.44 & 52.50 & 50.05 & 50.11 & 55.25 & 18.88 & 58.90 \\
        % BLOOMZ 1.7B & 59.29 & 51.98 & 51.66 & 54.58 & 57.05 & 28.53 & 57.90 \\
        % BLOOMZ 3B & 60.26 & 52.56 & 50.66 & 59.18 & 57.67 & 33.25 & 60.00 \\
        BLOOMZ 7.1B & 63.83 & 52.50 & 50.96 & 67.14 & 60.26 & 28.66 & 65.12 \\
        % mT0 Small & 51.57 & 52.16 & 51.39 & 38.63 & 46.85 & 18.37 & 50.71 \\
        % mT0 Base & 53.96 & 52.53 & 50.67 & 53.95 & 48.59 & 13.57 & 52.07 \\
        % mT0 Large & 56.24 & 53.63 & 52.03 & 54.26 & 49.47 & 19.78 & 54.35 \\
        % mT0 XL & 60.23 & 53.65 & 52.88 & 70.59 & 56.52 & 38.55 & 55.23 \\
        mT0 XXL & 64.79 & 55.34 & 54.59 & \textbf{86.79} & \textbf{64.30} & 39.85 & 58.92 \\
        Bactrian-X & 61.57 & 52.21 & 50.67 & 52.62 & 52.62 & 18.83 & 65.70 \\
        % LLaMA2 7B Chat & 54.12 & 52.70 & 49.30 & 51.92 & 48.44 & 29.52 & 61.57 \\
        LLaMA2 13B* & 56.94 & 51.01 & 48.97 & 43.15 & 49.61 & 34.98 & 65.05 \\
        \midrule
        \multicolumn{8}{c}{\textit{Southeast Asian LLM}}  \\
        \midrule
        SEALION 7B* & 59.71 & 51.68 & 48.65 & 34.00 & 55.21 & 21.92 & 63.79 \\
        SeaLLM 7B* & 64.24 & 53.46 & 49.72 & 58.54 & 55.11 & 33.60 & 68.55 \\
        \midrule
        \multicolumn{8}{c}{\textit{Indonesian LLM}}  \\
        \midrule
        Bactrian-Id & \textbf{65.34} & 51.79 & 48.48 & 45.84 & 56.27 & 22.95 & \textbf{69.14 }\\
        Merak 7B v4 & 62.30 & 52.46 & 50.65 & 77.78 & 55.82 & \textbf{46.27 }& 66.78 \\
        \midrule
        \multicolumn{8}{c}{\textit{Cendol}}  \\
        \midrule
        mT5 Small & 53.42 & 53.95 & 50.50 & 50.75 & 48.67 & 13.07 & 50.31 \\
        mT5 Base & 54.58 & 53.67 & 51.23 & 48.99 & 48.86 & 14.62 & 52.50 \\
        mT5 Large & 56.49 & 54.93 & 52.49 & 46.45 & 49.21 & 14.80 & 55.60 \\
        mT5 XL & 57.31 & 54.26 & 53.80 & 35.35 & 50.07 & 16.36 & 55.64 \\
        mT5 XXL & 62.30 & \textbf{55.80} & 52.71 & 43.02 & 53.95 & 14.46 & 56.87 \\
        LLaMA2 7B & 58.19 & 52.74 & \textbf{55.46} & 40.82 & 50.33 & 23.54 & 57.41 \\
        LLaMA2 13B & 56.82 & 52.21 & 54.08 & 37.32 & 52.52 & 21.87 & 59.09 \\
        \bottomrule
        \end{tabular}
    }
    \caption{Comparison of Cendol against various LLMs on local knowledge and cultural commonsense tasks. *We use the instruction-tuned versions.
    }
    \label{tab:cultural-eval}
\end{table}

We evaluate local knowledge using 7 cultural and local knowledge tasks. The results are presented in Table~\ref{tab:cultural-eval}. Our Cendol models are out-competed by some existing LLMs on all Indonesian language tasks, especially on IndoMMLU and IndoStoryCloze, where Cendol models perform the worst among all LLMs. Nonetheless, the best Cendol models achieve state-of-the-art performance on two local language tasks, i.e., MABL-jv and MABL-su. This highlights the existing multilingual, Southeast Asian, and Indonesian LLMs' limited understanding of Indonesian local languages. Furthermore, we observed a huge variance over different LLMs in some tasks, such as MAPS and IndoMMLU, which raises the question of whether some LLMs have seen the corresponding evaluation datasets.

% TODO: Instead of RLHF we do value transfer and we show that it works to osme extend .... Upper bound local languages LLMs, and we can achieve this much, and using a real RLHF can be further improve.

% \subsection{Safety Evaluation}  \sam{Appendix}
% \dummy{\lipsum[1-2]}

\subsection{Parameter Efficient Tuning Is Ineffective}
\label{sec:parameter-efficient}

% \todo{samcah}
We compare the effectiveness and efficiency of the parameter-efficient tuning method with LoRA~\cite{hu2022lora} with fully fine-tuned models with a similar training throughput. Specifically, we compare the LoRA-based Cendol mT5$_{XXL}$ model with two other models, i.e., Cendol mT5$_{large}$ and Cendol mT5$_{XL}$. As shown in Table~\ref{tab:parameter-efficient}, the training throughput of Cendol mT5$_{XXL}$ is $\sim$1.5x higher than Cendol mT5$_{XL}$ and is $\sim$0.34x lower than Cendol mT5$_{large}$. Nonetheless, in terms of other efficiency aspects, such as inference throughput and storage size, Cendol mT5$_{XXL}$ is less efficient than other models. In terms of quality, the training and evaluation losses of Cendol mT5$_{XXL}$ are much higher which leads to a worse performance downstream task performance for both NLU and NLG. These results demonstrate that despite reducing the computational resources compared to fully fine-tuned models of the same size, parameter-efficient tuning methods are less effective and less efficient compared to the smaller fully fine-tuned models in the case of language adaptation. We provide the Pareto-efficiency curve of Cendol mT5$_{XXL}$ compared to the other fully fine-tuned Cendol mT5 models in Appendix~\ref{app:pareto}. As an alternative solution for language adaptation, we introduce an alternative approach for improving the modeling efficiency through vocabulary adaptation~\cite{chau-etal-2020-parsing,poerner-etal-2020-inexpensive,tai-etal-2020-exbert,koto-etal-2021-indobertweet}.

\begin{table}[!t]
    \centering
    \resizebox{\linewidth}{!}{
        \begin{tabular}{l|c|c|c}
            \toprule
             \multirow{2}{*}{Aspect} & 
             \begin{tabular}{@{}c@{}}Cendol\\mT5$_{large}$\end{tabular} &
             \begin{tabular}{@{}c@{}}Cendol\\mT5$_{XL}$\end{tabular} & 
             \begin{tabular}{@{}c@{}} Cendol mT5$_{XXL}$\\(LoRA r=128) \end{tabular} \\
            \midrule
            Train Throughput ($\uparrow$) & \textbf{120} & 28 & 41 \\
            Eval Throughput ($\uparrow$) & \textbf{299} & 85 & 75 \\
            Parameter Size ($\downarrow$) & \textbf{1.2B} & 3.7B & 13B \\
            Train Loss ($\downarrow$) & 0.5819 & \textbf{0.2898} & 0.8015 \\
            Eval Loss ($\downarrow$) & 0.5938 & \textbf{0.2991} & 0.7715 \\
            Storage Size ($\downarrow$) & \textbf{4.8GB} & 14.8GB & 52GB \\
            \midrule
            NLU Perf. ($\uparrow$) & 52.07 & \textbf{59.77} & 47.47 \\
            NLG Perf. ($\uparrow$) & 35.21 & \textbf{ 42.76} & 32.09 \\
            \bottomrule
        \end{tabular}
    }
    \caption{Performance efficiency comparison of smaller fully fine-tuned and parameter-efficient tuning LLMs.}
    \label{tab:parameter-efficient}
\end{table}

\begin{table}[!t]
    \centering
    \resizebox{\linewidth}{!}{
        \begin{tabular}{c|c|c|c}
             \textbf{Factor} & \textbf{Vocab$\bm{^{ind}}$} & \textbf{Vocab$\bm{^{orig}}$} & $\bm{\Delta}$\textbf{Perf}.  \\
            \midrule
            \multicolumn{4}{c}{\textit{Model Efficiency}}\\
            \midrule
            Token Efficiency (Ind) &  46.34 tokens & 58.87 tokens & \cellcolor{lime}{$\uparrow$21.28\%} \\
            Token Efficiency (Oth) &  52.61 tokens & 61.74 tokens & \cellcolor{lime}{$\uparrow$14.79\%} \\
            Training (per batch) &  3.14s & 3.55s & \cellcolor{lime}{$\uparrow$11.50\%} \\
            Inference (per 100 steps) & 6.63s & 8.15s & \cellcolor{lime}{$\uparrow$18.71\%} \\
            \midrule
            \multicolumn{4}{c}{\textit{Downstream Performance}} \\
            \midrule
            NLU Performance & 58.51 & 55.4 & \cellcolor{lime}{$\uparrow$5.61\%} \\
            NLG Performance & 45.27 & 45.79 & \cellcolor{red}{\textcolor{white}{$\downarrow$1.14\%}} \\
            \bottomrule
        \end{tabular}
    }
    \caption{Efficiency and downstream tasks performance comparison between LLaMA2-7b with Indonesian-adapted vocabulary (Vocab$^{ind}$) and the LLaMA2-7b with original vocabulary (Vocab$^{orig}$).}
    \label{tab:token-efficiency}
\end{table}

Subword tokenization in LLMs generally produces longer sequences for low-resource language~\cite{ahia-etal-2023-languages} which makes it less efficient. By performing vocabulary adaptation~\cite{chau-etal-2020-parsing,tai-etal-2020-exbert,poerner-etal-2020-inexpensive}, we can improve the efficiency during the instruction tuning phase. Specifically, we use a subword vocabulary developed from Indonesian corpora, with the embedding initialized through averaging~\cite{koto-etal-2021-indobertweet}, and perform the first phase instruction tuning.
% (more detail in Appendix~\ref{app:token-efficiency}). 
As shown in Table~\ref{tab:token-efficiency}, the vocabulary adapted model steadily improve the token efficiency by 21.28\% for Indonesian text and 14.79\% for other local language text. This token efficiency results in improvements in both training and inference with around $\sim$11.50\% and  $\sim$18.71\% efficiency improvement, respectively. Additionally, in terms of downstream performance, the vocabulary-adapted model yields a similar performance compared to the model with the original vocabulary, with $5\%$ improvement on the NLU tasks and $1.14$\% reduction on the NLG tasks. Our result suggests that vocabulary adaptation with subword averaging provides an adequately representative initialization resulting in a significantly better efficiency and similar downstream task performance after the instruction tuning phase.

% in Appendix~\ref{app:token-efficiency}.
\subsection{Safety Transferability}
We conduct safety evaluations on truthfulness and harmful responses.
% , using 3 datasets, i.e., TruthfulQA~\cite{lin-etal-2022-truthfulqa}, ToxiGen~\cite{hartvigsen-etal-2022-toxigen}, and ImplicitHate~\cite{elsherief-etal-2021-latent}. 
The experiment and results are detailed in Appendix~\ref{app:safety}. It shows that Cendol is on par with other existing multilingual and regional LLMs in terms of safety. Interestingly, LLaMA2-based Cendol yields a much better safety score than mT5-based Cendol, suggesting the transferability of LLaMA2's safety pre-training to Cendol.

\section{Conclusion}

We introduce Cendol, a collection of Indonesian LLMs covering both decoder-only and encoder-decoder architecture over various model sizes, and Cendol Collection, a large-scale instruction-tuning dataset for Indonesian and its local languages. We highlight the effectiveness of Cendol on a wide range of tasks, achieving $\sim$20\% improvement for both NLU and NLG tasks. Cendol also generalizes to the local languages in Indonesia. Furthermore, we demonstrate our effort for human alignment through a supervised fine-tuning approach, which yields a significant improvement in terms of human favorability. We also discuss two limitations of Cendol: 1) human-preferred response needs to be further enhanced with a better human alignment approach, and 2) despite their amazing performance on NLP tasks, Cendol models still fall behind on capturing local knowledge and cultural values in Indonesia. Moreover, we analyze the generalization of Cendol to unseen tasks and languages, the ineffectiveness and inefficiency of LoRA for language adaptation, vocabulary adaptation as an efficient tuning alternative, and the safety of Cendol models.

\section*{Acknowledgements}

We thank Direktorat Jenderal Pendidikan Tinggi, Riset, dan Teknologi Kementerian Pendidikan, Kebudayaan, Riset, dan Teknologi (Ditjen DIKTI), Indonesia for partially supporting the computing resources used in this project. This work is also supported by the National Research Foundation, Singapore under its AI Singapore Programme; PhD Fellowship Award, the Hong Kong University of Science and Technology; and PF20-43679 Hong Kong PhD Fellowship Scheme, Research Grant Council, Hong Kong.

\section*{Limitations}

\paragraph{Better Human Alignment through Reinforcement Learning}
% \pascale{this is a huge limitation as governments are likely to ban such unsafe models in the future. So you need to talk about value transfer from other languages. Did you do it? }
One limitation of our work is the limited exploration of human value alignment. While our model can generate human-like text, it cannot generate responses that are aligned with human values, goals, and preferences. This lack of explicit human value alignment may result in the model producing outputs that are not only irrelevant but also potentially harmful or offensive to humans. Furthermore, without proper alignment, the model may not be able to understand and respond appropriately to the nuances and complexities of human communication, leading to misunderstandings and misinterpretations. Therefore, future work should focus on exploring and integrating human alignment techniques to ensure that the LLM can be safely and effectively used in real-world applications that involve human interaction.

\paragraph{Capturing Local Knowledge}

Another notable limitation that has emerged is the insufficient capability of our Cendol models to capture and reflect local cultural values accurately. This shortfall is partly due to the underrepresentation of diverse cultural contexts within the datasets used to train these LLMs. The majority of data feeding into the development of LLMs tends to be sourced from dominant languages and cultures, often overlooking the rich and nuanced expressions found in less-represented communities. Consequently, LLMs may exhibit biases that favor certain cultural norms and idioms, leading to misinterpretations or inappropriate responses when dealing with languages or dialects that are embedded with local cultural significance. The lack of cultural sensitivity in LLMs not only hinders effective communication but can also perpetuate stereotypes and misunderstandings.
% , potentially leading to cultural insensitivity or offensive responses.

\paragraph{Safety Evaluation}

Despite being the first to evaluate safety in the Indonesian, our current safety evaluation is done through translating the existing English safety corpora, i.e., TruthfulQA, ToxiGen, and ImplicitHate. Although most of the sentences remain valid, some of them are not natural and less culturally relevant to the regional context of Indonesia. These translated corpora are likely to miss important features such as local and cultural nuances, as well as contextual language which can hinder the effectiveness of the safety evaluation.  To make the safety evaluation more culturally relevant to the Indonesian context, the evaluation should utilize locally sourced Indonesian safety corpora. We expect future work to explore this direction to help ensure the safety evaluation is sensitive to the local cultural and social environment and provide more accurate insights into potential safety risks specific to the regional values.

\paragraph{Single-Turn Human-Computer Interaction} 

Although our instruction-tuning data and user-oriented evaluation primarily focus on building general-purpose LLMs which are commonly expected to be able to respond interactively in a multi-turn manner. It is essential to acknowledge that our Cendol$^{inst}$ and Cendol$^{chat}$ models are not currently optimized for handling multi-turn dialogues. In other words, they are not expected to be able to engage in a continuous human-computer interaction. This can result in less coherent and less effective responses when compared to models specifically designed for a continuous human-computer interaction. Therefore, future work should focus more on developing a multi-turn dialogue system, by preserving the context and the interactions between the user and the model in previous turns and carry them to future turns. 

\section*{Ethics Statement}

Our research underscores the imperative of democratizing access to NLP technology for underrepresented languages, with a particular emphasis on Indonesian and its local languages. We recognize and embrace the ethical responsibilities inherent in language research, acutely aware of its potential impact on diverse linguistic communities. Our commitment to inclusivity, cultural relevance, and fairness is the cornerstone of our study. Transparent and equitable collaboration is the lifeblood of our work, and we uphold a fair and transparent scoring guideline that aligns with our core principles.

Throughout our study, we have made conscious efforts to engage with language communities, involve local experts, and respect their linguistic and cultural nuances. This effort is not merely a component of our research - it is an ongoing dialogue, fostering mutual respect and understanding. Our ultimate goal is to contribute to a more inclusive NLP landscape, one that celebrates linguistic diversity and mitigates biases. By encouraging further collaboration and ensuring that the voices of underrepresented language communities are heard, we aim to address their specific needs in the development of language technology. 
% We continue to strive towards an NLP future that is accessible and equitable for every language, no matter how underrepresented it may be.

\bibliography{anthology,custom}
\bibliographystyle{acl_natbib}

\newpage

\appendix

% \section{Tasks and Datasets in Cendol Collection}
% \label{app:cendol-collection}

% We list out all the tasks incorporated in Cendol Collection in Table~\ref{}. 
% Note that, Add paragraph for discussing about datasets licenses 

\section{Tuning Hyperparameters}
\label{app:hyperparameters}

We provide detailed hyperparameter tuning for developing Cendol$^{inst}$ and Chat for each different model architecture in Table~\ref{tab:hyperparams}.

\begin{table}[!h]
    \centering
    \resizebox{\linewidth}{!}{
    \begin{tabular}{c|c|c|c|c}
    \toprule
        Hyperparameter & mT5 & mT5 & LLaMA-2 & LLaMA-2 \\
         & small..xl & xxl & 7B & 13B \\
    \midrule
        \multicolumn{5}{c}{Cendol$^{inst}$} \\
    \midrule
        max\_input\_length  & 512 & 512 & 512 & 512 \\
        max\_output\_length & 256 & 256 & 768 & 768 \\
        batch\_size & 128 & 128 & 128 & 128 \\
        bfp16 & True & True & True & True \\
        zero\_config & zero-3 & zero-3 & zero-3 & zero-3 \\
        lr  & 3e-4 & 2e-4 & 2e-5 & 2e-4 \\
        lora\_r & - & 128 & - & 128 \\
        lora\_alpha & - & 128 & - & 128 \\
        lora\_dropout & - & 0.05 & - & 0.05 \\
    \midrule
        \multicolumn{5}{c}{Cendol$^{chat}$} \\
    \midrule
        max\_input\_length  & 512 & 512 & 512 & 512 \\
        max\_output\_length & 256 & 256 & 768 & 768 \\
        batch\_size & 128 & 128 & 128 & 128 \\
        bfp16 & True & True & True & True \\
        zero\_config & zero-3 & zero-3 & zero-3 & zero-3 \\
        lr  & 3e-5 & 1e-4 & 1e-5 & 1e-4 \\
        lora\_r & - & 128 & - & 128 \\
        lora\_alpha & - & 128 & - & 128 \\
        lora\_dropout & - & 0.05 & - & 0.05 \\
    \bottomrule
    \end{tabular}
    }
    \caption{List of hyperparameter settings used during the instruction-tuning of Cendol$^{inst}$ and Cendol$^{chat}$.}
    \label{tab:hyperparams}
\end{table}

\section{Safety Evaluation} 
\label{app:safety}

\begin{table}[!h]
\centering
\resizebox{0.7\linewidth}{!}{%
\begin{tabular}{l|l|r}
\toprule
\textbf{model} & \textbf{lang} & \textbf{accuracy} \\
\midrule
Bactrian-X & ind & 67.26\% \\
Bactrian-Id & ind & \textbf{98.41\%} \\
% BLOOMZ 7.1B & ind & 44.12\% \\
% mT0 XXL & ind & 84.58\% \\
LLaMA2 7B Chat & ind & 95.72\% \\
LLaMA2 13B Chat & ind & 97.28\% \\
\midrule
SEALION 7B Instruct-nc & ind & 53.46\% \\
SeaLLM 7B Chat & ind & 76.90\% \\
Merak 7B v4 & ind & \textbf{88.98\%} \\
\midrule
Cendol mT5 Small & ind & 47.25\% \\
Cendol mT5 Base & ind & 87.76\% \\
Cendol mT5 Large & ind & 98.56\% \\
Cendol mT5 XL & ind & 95.32\% \\
Cendol mT5 XXL & ind & \underline{\textbf{98.75\%}} \\
Cendol LLaMA2 7B & ind & 31.21\% \\
Cendol LLaMA2 13B & ind & 71.63\% \\
\bottomrule
\end{tabular}%
}
\caption{Evaluation of Cendol and benchmark LLMs on the automatic truthful benchmark (Higher is better). 
% In this table, we present the percentage of generations that are both truthful and informative (the higher the better). 
The overall most truthful model is denoted by \underline{underline}, while within the group, they are denoted by \textbf{bold}.}
\label{tab:safety_truthful}
\end{table}

\begin{table*}[!t]
\centering
\resizebox{\textwidth}{!}{%
\begin{tabular}{l|r|r|r|r|r|r|r|r|r|r|r|r|r|r}
\toprule
\textbf{model} & \textbf{asian} & \textbf{black} & \textbf{chinese} & \textbf{jewish} & \textbf{latino} & \textbf{lgbtq} & \begin{tabular}[c]{@{}c@{}}\textbf{mental}\\\textbf{disability}\end{tabular} & \textbf{mexican} & \begin{tabular}[c]{@{}c@{}}\textbf{middle}\\\textbf{eastern}\end{tabular} & \textbf{muslim} & \begin{tabular}[c]{@{}c@{}}\textbf{native}\\\textbf{american}\end{tabular} & \begin{tabular}[c]{@{}c@{}}\textbf{physical}\\\textbf{disability}\end{tabular} & \textbf{women} & \textbf{avg} \\
\midrule
Bactrian-X & 37.34\% & 31.96\% & 38.47\% & 36.74\% & 29.52\% & 31.49\% & 24.94\% & 29.96\% & 27.29\% & 30.93\% & \textbf{24.89\%} & 23.06\% & 23.76\% & 30.03\% \\
Bactrian-Id & \underline{\textbf{37.54\%}} & \underline{\textbf{33.45\%}} & \underline{\textbf{38.51\%}} & \underline{\textbf{37.11\%}} & \underline{\textbf{30.75\%}} & \underline{\textbf{31.99\%}} & \textbf{25.07\%} & \underline{\textbf{31.16\%}} & \textbf{27.37\%} & \textbf{31.25\%} & 23.99\% & \textbf{23.31\%} & \textbf{24.63\%} & \underline{\textbf{30.47\%}} \\
% mT0 XXL & 4.87\% & 4.86\% & 4.82\% & 4.23\% & 3.07\% & 3.14\% & 3.95\% & 3.32\% & 3.03\% & 3.07\% & 1.98\% & 3.02\% & 2.32\% & 3.51\% \\
LLaMA2 7B Chat & 29.25\% & 26.09\% & 28.01\% & 23.73\% & 16.55\% & 21.79\% & 15.80\% & 18.19\% & 19.96\% & 23.76\% & 21.08\% & 16.12\% & 16.19\% & 21.27\% \\
LLaMA2 13B Chat & 27.07\% & 24.92\% & 25.40\% & 23.08\% & 16.15\% & 20.96\% & 14.88\% & 17.33\% & 18.46\% & 23.10\% & 21.46\% & 14.85\% & 15.51\% & 20.24\% \\
\midrule
SEALION 7B Instruct & \textbf{36.61\%} & 32.62\% & \textbf{37.02\%} & \textbf{30.99\%} & \textbf{22.91\%} & \textbf{30.74\%} & \textbf{25.96\%} & \textbf{22.46\%} & 23.63\% & 31.32\% & 26.08\% & \textbf{24.71\%} & \textbf{23.98\%} & \textbf{28.39\%} \\
SeaLLM 7B Chat & 34.50\% & \textbf{33.20\%} & 33.70\% & 30.49\% & 20.95\% & 24.95\% & 23.15\% & 22.25\% & \textbf{23.83\%} & \underline{\textbf{32.20\%}} & \textbf{27.58\%} & 21.99\% & 18.09\% & 26.68\% \\
Merak 7B v4 & 24.06\% & 20.91\% & 20.98\% & 20.81\% & 12.50\% & 17.12\% & 9.73\% & 13.62\% & 12.45\% & 19.86\% & 14.44\% & 9.28\% & 10.92\% & 15.90\% \\
\midrule
Cendol mT5 Small & 1.86\% & 1.10\% & 2.17\% & 1.22\% & 2.42\% & 1.30\% & 2.06\% & 2.36\% & 2.12\% & 1.31\% & 0.16\% & 2.15\% & 1.14\% & 1.64\% \\
% Cendol mT5 Small Chat & 5.36\% & 4.55\% & 4.57\% & 4.36\% & 3.60\% & 3.51\% & 3.50\% & 3.70\% & 2.58\% & 3.59\% & 1.79\% & 3.87\% & 1.47\% & 3.57\% \\
Cendol mT5 Base & 8.30\% & 3.88\% & 10.10\% & 9.57\% & 4.50\% & 7.23\% & 7.08\% & 3.36\% & 12.64\% & 7.50\% & 3.12\% & 3.32\% & 4.55\% & 6.55\% \\
% Cendol mT5 Base Chat & 9.88\% & 5.10\% & 12.05\% & 10.24\% & 4.94\% & 8.39\% & 6.49\% & 4.17\% & 12.68\% & 7.38\% & 2.46\% & 2.92\% & 4.03\% & 6.98\% \\
Cendol mT5 Large & 4.69\% & 3.39\% & 4.86\% & 3.40\% & 3.83\% & 3.37\% & 4.42\% & 3.33\% & 4.67\% & 4.80\% & 3.59\% & 5.16\% & 4.09\% & 4.12\% \\
% Cendol mT5 Large Chat & 4.71\% & 3.28\% & 5.11\% & 2.87\% & 2.53\% & 3.64\% & 2.44\% & 2.77\% & 2.19\% & 3.84\% & 1.48\% & 3.39\% & 1.87\% & 3.09\% \\
Cendol mT5 XL & 6.24\% & 4.31\% & 6.35\% & 5.66\% & 3.58\% & 4.60\% & 6.47\% & 3.83\% & 6.08\% & 7.93\% & 2.91\% & 8.87\% & 6.65\% & 5.65\% \\
% Cendol mT5 XL Chat & 3.44\% & 2.54\% & 3.71\% & 1.99\% & 2.21\% & 2.06\% & 1.50\% & 2.04\% & 1.50\% & 3.12\% & 0.80\% & 2.00\% & 1.63\% & 2.19\% \\
Cendol mT5 XXL & 3.09\% & 1.74\% & 2.64\% & 1.44\% & 1.48\% & 1.55\% & 2.23\% & 1.40\% & 1.95\% & 1.80\% & 1.00\% & 1.96\% & 1.42\% & 1.82\% \\
% Cendol mT5 XXL Chat & 9.62\% & 6.08\% & 9.10\% & 5.96\% & 4.41\% & 6.57\% & 5.11\% & 4.69\% & 4.33\% & 7.59\% & 3.95\% & 7.15\% & 2.92\% & 5.96\% \\
Cendol LLaMA2 7B & \textbf{30.83\%} & 31.64\% & \textbf{31.55\%} & 25.00\% & \textbf{28.82\%} & \textbf{30.49\%} & \underline{\textbf{28.07\%}} & \textbf{29.62\%} & \underline{\textbf{28.90\%}} & 27.50\% & \underline{\textbf{31.18\%}} & \underline{\textbf{25.10\%}} & \underline{\textbf{27.86\%}} & \textbf{28.97\%} \\
% Cendol LLaMA2 7B Chat & \textbf{35.96\%} & 30.84\% & \textbf{35.10\%} & \textbf{29.12\%} & 19.85\% & 28.72\% & 20.00\% & 22.21\% & 26.85\% & \underline{\textbf{33.63\%}} & 23.42\% & 20.01\% & 22.52\% & 26.79\% \\
Cendol LLaMA2 13B & 29.23\% & \textbf{32.62\%} & 26.51\% & \textbf{27.72\%} & 21.19\% & 24.85\% & 22.45\% & 19.41\% & 26.66\% & \textbf{31.85\%} & 23.86\% & 18.33\% & 19.01\% & 24.90\% \\
% Cendol LLaMA2 13B Chat & 25.56\% & 21.52\% & 23.71\% & 21.54\% & 13.62\% & 17.95\% & 11.63\% & 14.62\% & 14.84\% & 20.38\% & 16.26\% & 11.47\% & 11.97\% & 17.31\%\\
\bottomrule
\end{tabular}%
}
\caption{Evaluation of Cendol and benchmark LLMs on ToxiGen (higher means less toxic). 
% In this table, we present the percentage of safety score, where healthy models will score 1, in which the harmful sentences have less probability to be produced by the models than the benign sentences, and vice versa. 
The overall least toxic models are denoted by \underline{underline}, while within the group, they are denoted by \textbf{bold}.}
\label{tab:safety_toxigen}
\end{table*}

In our analysis, we focus on assessing the language model's performance in terms of its \textbf{truthfulness} and \textbf{toxicity}. Specifically, \textbf{truthfulness} pertains to the model's ability to avoid disseminating information that is inaccurate due to misconceptions or erroneous beliefs. Meanwhile, \textbf{toxicity} refers to the model's propensity to produce content that is toxic, rude, adversarial, or implicitly hateful. We leverage three distinct datasets: TruthfulQA~\cite{lin-etal-2022-truthfulqa}, which benchmarks the veracity of responses, ImplicitHate~\cite{elsherief-etal-2021-latent}, designed to identify underlying hate speech, and ToxiGen, a resource for gauging the generation of toxic text~\cite{hartvigsen-etal-2022-toxigen}. 

We first translate the dataset using a distilled NLLB model with 1.3B parameters~\footnote{\url{facebook/nllb-200-distilled-1.3B}} and evaluate all the datasets on a set of 31 LLMs including Cendol and multilingual, regional, and Indonesian-adapted LLMs. For TruthfulQA, we report the accuracy score on the MC1 subset as the percentage of generations that are both truthful and informative. For ToxiGen and ImplicitHate, we employ an automatic safety score evaluation metric from \citet{hosseini-etal-2023-empirical} as the percentage of likeliness of the model producing benign over harmful sentences. We show the evaluation result on TruthfulQA, ToxiGen, and ImplicitHate, in Table~\ref{tab:safety_truthful}, Table~\ref{tab:safety_implicithate}, and  Table~\ref{tab:safety_toxigen}, respectively.

\begin{table}[!t]
\centering
\resizebox{0.5\columnwidth}{!}{%
\begin{tabular}{l|r}
\toprule
\textbf{model} & \textbf{score} \\
\midrule
Bactrian-X & 29.62\% \\
Bactrian-Id & \underline{\textbf{30.46\%}} \\
% BLOOMZ 7.1B & \underline{\textbf{43.45\%}} \\
% mT0 XXL & 3.14\% \\
LLaMA2 7B Chat & 20.18\% \\
LLaMA2 13B Chat & 19.44\% \\
\midrule
SEALION 7B Instruct & \textbf{27.57\%} \\
SeaLLM 7B Chat & 26.23\% \\
Merak 7B v4 & 18.10\% \\
\midrule
Cendol mT5 Small & 1.18\% \\
% Cendol mT5 Small Chat & 4.39\% \\
Cendol mT5 Base & 5.45\% \\
% Cendol mT5 Base Chat & 6.51\% \\
Cendol mT5 Large & 3.68\% \\
% Cendol mT5 Large Chat & 2.52\% \\
Cendol mT5 XL & 4.39\% \\
% Cendol mT5 XL Chat & 1.96\% \\
Cendol mT5 XXL & 2.19\% \\
% Cendol mT5 XXL Chat & 9.42\% \\
Cendol LLaMA2 7B & \textbf{25.30\%} \\
% Cendol LLaMA2 7B Chat & \textbf{25.76\%} \\
Cendol LLaMA2 13B & 22.20\% \\
% Cendol LLaMA2 13B Chat & 18.73\% \\
\bottomrule
\end{tabular}%
}
\caption{Evaluation of Cendol and benchmark LLMs on ImplicitHate (higher means less hateful). 
% In this table, we present the percentage of safety score, where healthy models will score 1, in which the implicitly hateful sentences have less probability to be produced by the models than the benign sentences, and vice versa. 
The overall least harmful model is denoted by \underline{underline}, while within the group, they are denoted by \textbf{bold}.}
\label{tab:safety_implicithate}
\end{table}

\paragraph{Safety Evaluation Result} Cendol mT5 XXL model outperforms mT0 XXL, demonstrating a 16.75\% enhancement in both truthfulness and informativeness. Additionally, evaluations of the Cendol mT5 XXL Chat indicate a reduction in implicit hate speech by 6.28\% points and a decrease in toxicity by 2.45\% points. A trend is discernible within the Cendol mT5 model series, revealing that an increase in model size correlates with improvements in truthfulness and informativeness. Although the Cendol mT5 series excels in these areas, the Cendol LLaMA2 series demonstrates a superior capability in generating significantly lower levels of toxic and hateful outputs. Interestingly, LLaMA2~\cite{touvron2023llama2} incorporates safety measures during the pretraining, which is lacking in mT5~\cite{xue-etal-2021-mt5}. \textbf{This suggests that there is a transfer of safety across languages from English to Indonesian}. In addition, Cendol models are also comparably more truthful and less toxic, when compared to the local, regional, and Indonesian-adapted models. For instance, the Cendol mT5-XL model, with 3.7 billion parameters, is found to be 6.34 percentage points more truthful than the Merak-7B-v4 model. In terms of toxicity, the Cendol LLaMA2 7B model is less toxic by 0.58 percentage points compared to its regional counterpart, the SEALION 7B Instruct model.  It's noteworthy, however, that regional models are generally less prone to producing implicit hate speech.

% \usepackage{graphicx}

% \section{Carbon Footprint}
% \label{app:carbon-footprint}
% \dummy{\lipsum[4]}

% \section{Evaluation Suite}
% \label{app:eval-suite}

% \begin{table}[!t]
%     \centering
%     \resizebox{\linewidth}{!}{
%     \begin{tabular}{l|c|c|c}
%         \toprule
%         Type & Task Type & Task & Dataset \\
%         \midrule
%         \multirow{7}{*}{NLU} & \multirow{3}{*}{Seen Tasks} & Entailment Classification & Wrete \\
%          &  & Sentiment Classification & ID Google Play Review, Code Mixed jv id \\
%          &  & Hate Speech Detection & id\_hsd\_nofaaulia \\
%          \cline{2-5}
%          & \multirow{4}{*}{Unseen Tasks} & Answer Grading & id\_short\_answer\_grading\_nusantara\_pairs\_score', \\
%          &  & Stance Detection & id\_stance\_nusantara\_pairs', \\
%          &  & Next Tweet Prediction & indolem\_ntp\_nusantara\_pairs', \\
%          &  & Dialect Prediction & jadi\_ide\_nusantara\_text', \\
%          \midrule
%         \multirow{3}{*}{NLG} & \multirow{3}{*}{Seen Tasks} & Machine Translation & Flores 200 \\
%          &  & Summarization & xl sum \\
%          &  & Paraphrasing & stif indonesia \\
%         \bottomrule
%     \end{tabular}
%     }
%     \caption{Evaluation tasks details.}
%     \label{tab:task_details}
% \end{table}

\section{Human Evaluation Guidelines}
\label{app:human-eval}

\begin{figure}[!h]
    \centering
    \resizebox{\linewidth}{!}{
    \includegraphics{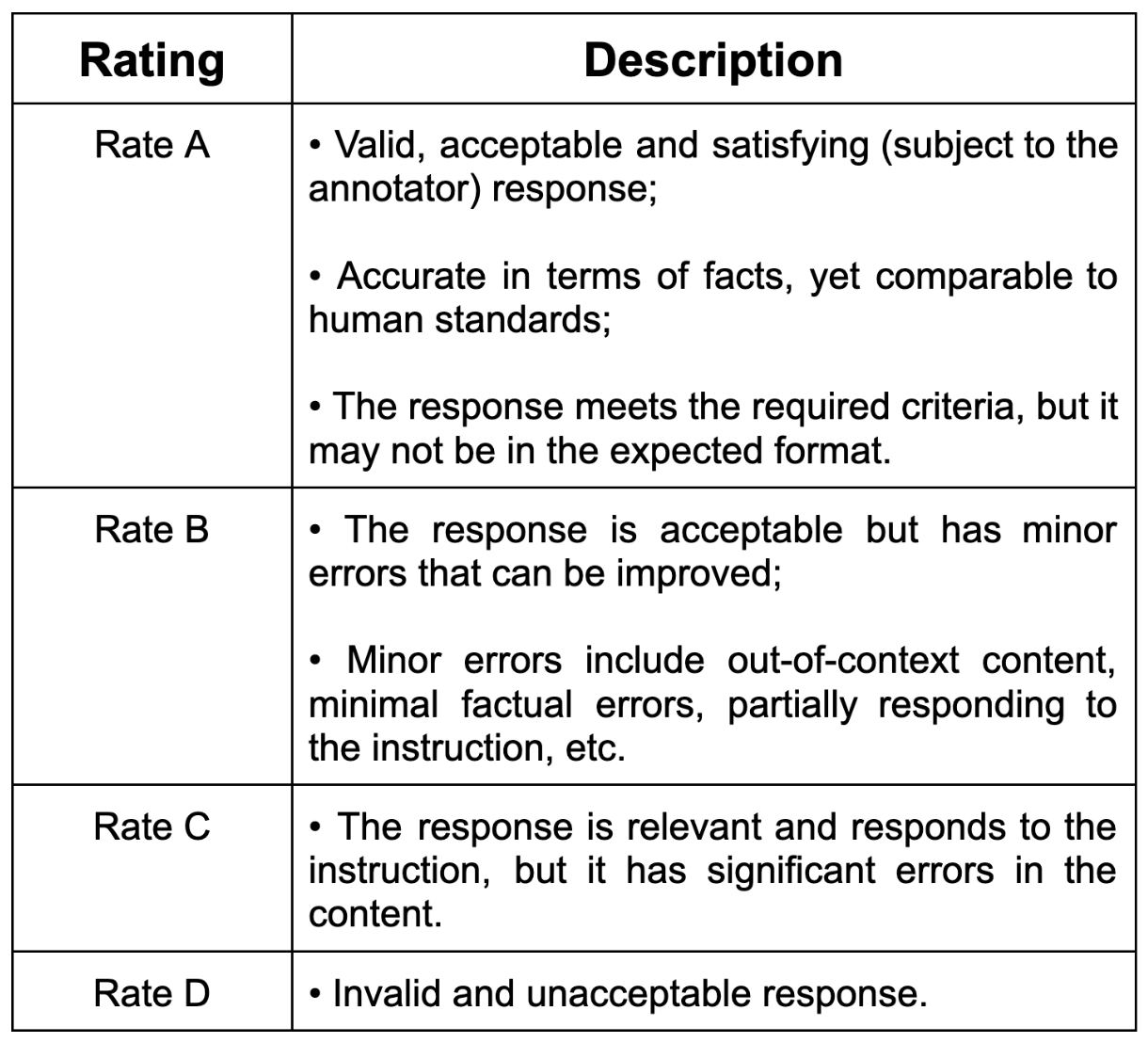}
    }
    \caption{Human annotation guideline that is incorporated in our human evaluation of Cendol.}
    \label{fig:annot-guide}
\end{figure}

We adopt the human evaluation and annotation guidelines from prior works~\cite{wu2023laminilm,li2023bactrianx}. We provide the detailed human evaluation guideline in Figure~\ref{fig:annot-guide}.

\section{Pareto-Efficiency of Cendol Models}
\label{app:pareto}

\begin{figure}[!h]
    \centering
    \resizebox{0.87\linewidth}{!}{
    \includegraphics{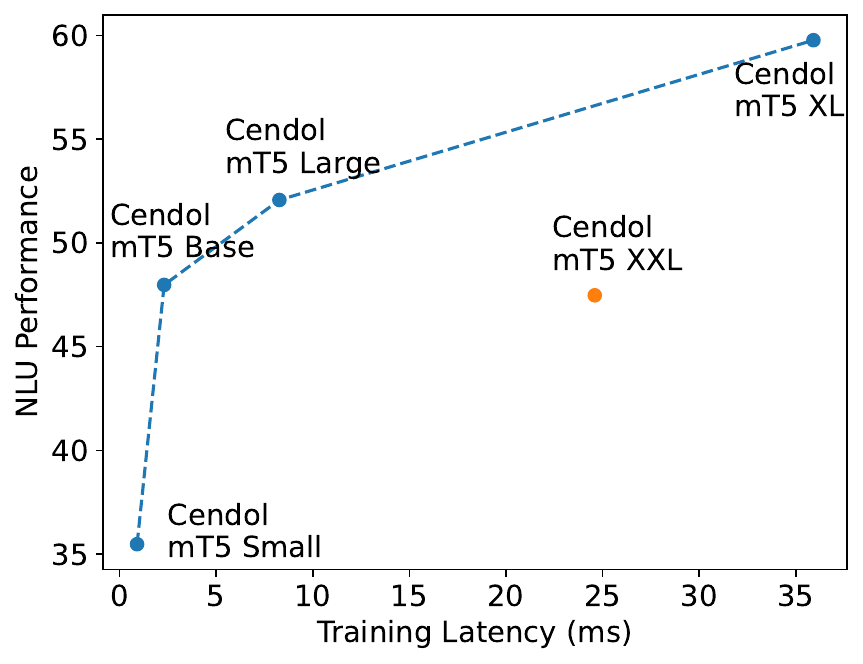}
    }
    \caption{Pareto-efficiency curve of Cendol mT5 models. Parameter efficient method leads to a non-pareto optimal point as shown in the case of Cendol mT5$_{XXL}$.}
    \label{fig:pareto}
\end{figure}

We showcase the Pareto-efficiency curve of the Cendol mT5 models in Figure~\ref{fig:pareto}.

\end{document}